\documentclass{article}

\PassOptionsToPackage{numbers, compress}{natbib}

\usepackage[main, final]{neurips_2025}

\usepackage[utf8]{inputenc} % allow utf-8 input
\usepackage[T1]{fontenc}    % use 8-bit T1 fonts
\usepackage[colorlinks]{hyperref}       % hyperlinks
\usepackage{url}            % simple URL typesetting
\usepackage{booktabs}       % professional-quality tables
\usepackage{amsfonts}       % blackboard math symbols
\usepackage{nicefrac}       % compact symbols for 1/2, etc.
\usepackage{microtype}      % microtypography
\usepackage{xcolor}         % colors
\usepackage{cancel}
\usepackage{multirow}       % For multi-row cells
\usepackage{amssymb}
\usepackage{tablefootnote}
\usepackage{graphicx}
\usepackage[para,online,flushleft]{threeparttable}

\usepackage{amsmath,amsfonts,bm}

\newcommand{\figleft}{{\em (Left)}}

\newcommand{\figright}{{\em (Right)}}

\def\eqref#1{equation~\ref{#1}}

\def\1{\bm{1}}

\DeclareMathAlphabet{\mathsfit}{\encodingdefault}{\sfdefault}{m}{sl}
\SetMathAlphabet{\mathsfit}{bold}{\encodingdefault}{\sfdefault}{bx}{n}

\def\gQ{{\mathcal{Q}}}

\newcommand{\E}{\mathbb{E}}

\DeclareMathOperator*{\argmax}{arg\,max}
\DeclareMathOperator*{\argmin}{arg\,min}

\usepackage[capitalise]{cleveref} % Ensure cleveref is loaded after amsmath
\usepackage{booktabs}
\usepackage{xcolor}
\usepackage{colortbl}
\usepackage{makecell}
\usepackage{wrapfig}
\usepackage{float}
\usepackage{subcaption}

\newcommand{\overbar}[1]{\mkern 1.5mu\overline{\mkern-1.5mu#1\mkern-1.5mu}\mkern 1.5mu}

\definecolor{mypurple}{HTML}{7A4988}
\definecolor{myorange}{HTML}{E77500}
\definecolor{myblue}{HTML}{598BE7}
\hypersetup{urlcolor=myblue,citecolor=myblue,linkcolor=myblue}

\title{Normalizing Flows are Capable Models for Continuous Control}

\author{%
  Raj Ghugare \qquad Benjamin Eysenbach \\
  Department of Computer Science\\
  Princeton University\\
  \texttt{rg9360@princeton.edu}
}

\begin{document}
\maketitle
\vspace{-2em}
\begin{abstract}
\vspace{-0.5em}
Modern reinforcement learning (RL) algorithms have found success by using probabilistic models, such as transformers, energy-based models, and diffusion/flow-based models. To this end, researchers often choose to pay the price of accommodating these models into their algorithms -- diffusion models are expressive, but are computationally intensive due to their reliance on solving differential equations, while autoregressive transformer models are scalable but typically require learning discrete representations. Normalizing flows (NFs), by contrast, seem to provide an appealing alternative, as they enable likelihoods and sampling without solving differential equations or autoregressive architectures. However, their potential in RL has received limited attention, partly due to the prevailing belief that normalizing flows lack sufficient expressivity. We show that this is not the case. Building on recent work in NFs, we propose a single NF architecture which integrates seamlessly into RL algorithms, serving as a policy, Q-function, and occupancy measure. Our approach leads to much simpler algorithms, and achieves higher performance in imitation learning, offline, goal conditioned RL and unsupervised RL.\footnote{The code for all experiments can be found here : \href{https://github.com/Princeton-RL/normalising-flows-4-reinforcement-learning}{https://github.com/Princeton-RL/normalising-flows-4-reinforcement-learning}.}
\looseness=-1
\end{abstract}

\begin{figure}[h]
    \centering
    \vspace{-1em}
    \includegraphics[width=1\linewidth]{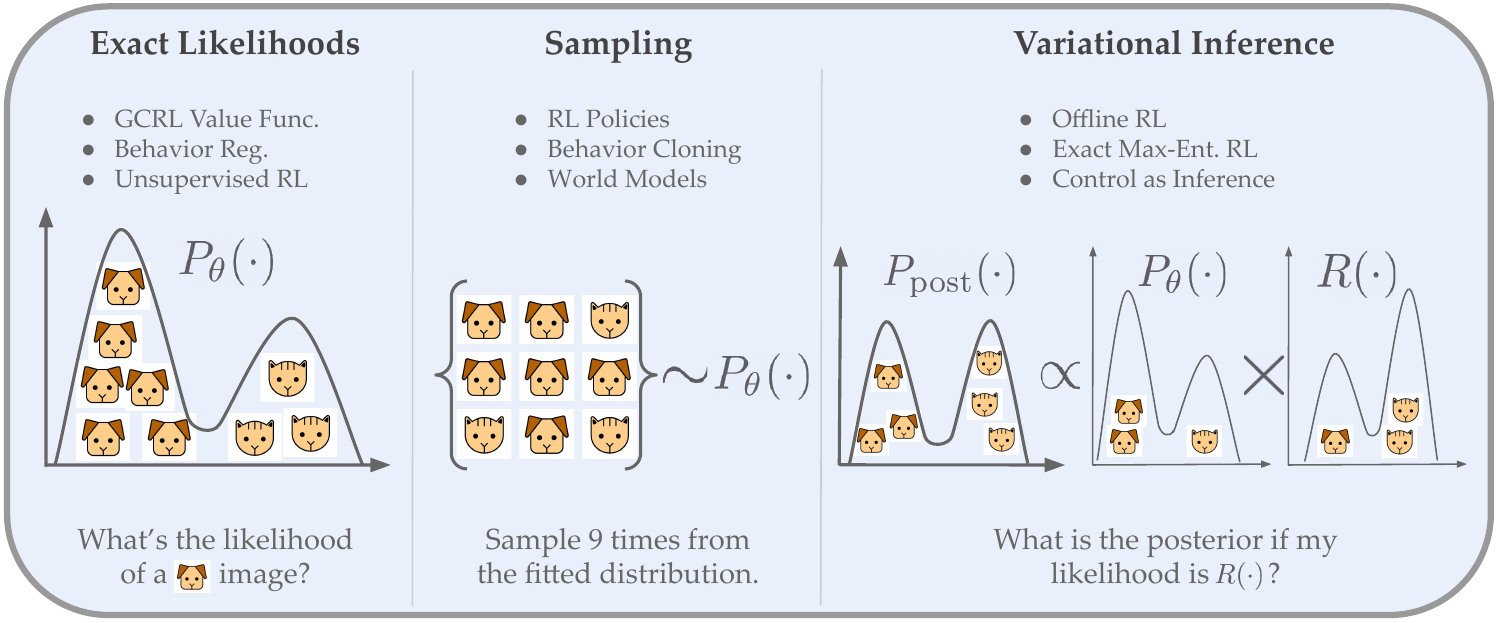}
    \vspace{-1em}
    \caption{\label{fig:nf-capabilities} \footnotesize \textbf{Probabilistic models and RL.} 
    Three capabilities of probabilistic models are important for RL: efficient likelihood computation, sampling, and compatibility with variational inference (VI).
    These capabilities are at the core of most RL algorithms~(See~\cref{sec:rl-vi-mle} for details). For example, behavior cloning requires maximizing likelihoods and sampling~\cite{Bain1995_bc}, GCRL value functions require estimating future discounted state probabilities~\cite{eysenbach2021clearning}, and MaxEnt RL and RL finetuning requires variational inference~\cite{levine2018cic}. However, models like diffusion and autoregressive transformers used in RL today support only a subset of these capabilities, making their application complex. In this paper, we show how normalizing flows, which posses these properties, can be a generally capable model family for RL.} 
\end{figure}

\begin{table}[h]
 \vspace{-2em}
    \centering
    \begin{threeparttable}
    \caption{\footnotesize \textbf{Capabilities of probabilistic models.} NFs have capabilities that make them applicable to a large number of RL problems~(\cref{sec:rl-vi-mle}), including the ability to compute exact likelihoods, sample from complex distributions, and be trained via both  maximum likelihood and variational inference~(\cref{sec:prelimns}). Despite these capabilities, research on NFs in RL has been limited compared to other state of the art models. Is this because NFs lack expressivity? Our experiments~(\cref{sec:experiments}) provide compelling evidence that this is not the case.} 
    \begin{tabular}{l c c c c}
        \toprule
        \rowcolor{white}  \text{Model family} & \text{Exact Likelihoods} & \text{Sampling} & \text{Variational Inference}\tnote{*}\\
        \midrule
        Variational Autoencoders & \cellcolor{red!20}Lower bound & \cellcolor{green!20}Yes & \cellcolor{yellow!20}No \\
        Generative Adversarial networks & \cellcolor{red!20}No & \cellcolor{green!20}Yes & \cellcolor{yellow!20}No \\
        Energy Based models & \cellcolor{yellow!20}Unnormalized & \cellcolor{yellow!20}MCMC & \cellcolor{yellow!20}No \\
        Diffusion models &  \cellcolor{yellow!20}Yes, via ODEs & \cellcolor{green!20}Unrolling SDEs & \cellcolor{yellow!20}No \\
        {\color{myorange}\texttt{Normalizing Flows}} & \cellcolor{green!20}Yes & \cellcolor{green!20}Yes & \cellcolor{green!20}Yes \\
        \bottomrule
        \end{tabular}
    \label{tab:nf-capabilities}
    \begin{tablenotes}
    {\item[*] \footnotesize Although directly performing VI~(posterior sampling and estimation) is generally intractible for the first four model families we have still marked them with yellow because additional algorithms for VI do exist~\cite{Blei_2017}.}
    \end{tablenotes}
    \end{threeparttable}
\end{table}
\vspace{-2em}
% \looseness=-1

\section{Introduction}
\label{sec:intro}

%What is the problem?
Reinforcement learning (RL) algorithms often require probabilistic modeling of high dimensional and complex distributions. For example, offline RL or behavior cloning (BC) from large scale robotics datasets requires learning a multi-modal policy~\cite{Bain1995_bc, openx, black2024pi0} and value function estimation in goal-conditioned RL (GCRL) requires estimating probabilities over future goals~\cite{eysenbach2021clearning}. Different algorithms require different capabilities, including likelihood estimation, variational inference (VI) and efficient sampling (see~\Cref{fig:nf-capabilities}). A model trained via BC must maximize action likelihoods during training and support fast action sampling for real-time control. Most offline or finetuning RL algorithms require policy models to be able to backpropagate gradients of the value function through sampled actions. For MaxEnt RL with continuous actions, models should be optimizable using variational inference. However, obtaining these abilities in current models often comes at the cost of compute and complexity~(\cref{tab:nf-capabilities}). For example, diffusion models~\cite{sohldickstein2015diff} are highly expressive yet computationally demanding: training, sampling, and evaluating log likelihoods all require solving a differential equation. Autoregressive transformer models~\cite{vaswani2023aiayn} are scalable and expressive, but are typically limited to discrete settings; using them for RL in continuous spaces requires learning discrete representations of actions~\cite{shafiullah2022cloningk} or states~\cite{micheli2023twm}. Gaussian models offer computational efficiency but lack expressivity. \textit{Can we design expressive probabilistic models with all the capabilities in~\cref{fig:nf-capabilities}?}
\looseness=-1

Normalizing Flows (NFs) are among the most flexible probabilistic models: they $(1)$ support both likelihood maximization and variational inference training, and $(2)$ enable efficient sampling and exact likelihood computation (see~\cref{tab:nf-capabilities}). Yet, NFs have received far less attention from the RL community, perhaps due to the (mis)conception that they have restricted architectures or training instabilities~\cite{papamakarios2021nfs}. This paper demonstrates that neither of these notions hold NFs back in practice: NFs can match the expressivity of diffusion and autoregressive models while being significantly simpler. Moreover, unlike other models, NFs do not require additional machinery to be integrated with existing RL frameworks, enabling simpler algorithms that are competitive with if not better than their counterparts.
\looseness=-1

Our primary contribution is to shed light on the effectiveness and NFs in RL. Building on prior work~\cite{kingma2018glow, dinh2017realnvp}, we develop a simple NF architecture and evaluate it across 82 tasks spanning five different RL settings, including offline imitation learning, offline RL, goal-conditioned RL, and unsupervised RL:
\begin{itemize}
    \item On offline imitation learning, NFs are competitive with baselines using state of the art generative models such as diffusion and transformers, while using significantly less compute~(6$\times$ fewer parameters than diffusion policies).
    \item On offline conditional imitation learning, NFs outperform BC baselines using models like flow matching, and offline RL algorithms which learn a value function, across 45 tasks.
    \item On offline RL problems, NFs are competitive with strong baselines using diffusion / flow matching models on most tasks, outperforming all of them on $15/30$ tasks.
    \item On unsupervised GCRL, NFs can rival contrastive learning based algorithms, despite making fewer assumptions.
\end{itemize}

\section{Related Work}
\label{sec:related}

\paragraph{Normalizing flows.} This work builds on the rich literature of normalizing flow architectures~\cite{dinh2015nice, rezende2016nf, kingma2018glow, kingma2017iafs, huang2018naf, papamakarios2018maf, zhai2024tarf}. NICE~\cite{dinh2015nice} introduce the idea of stacking neural function blocks that perform shifting, making density estimation efficient. RealNVP~\cite{dinh2017realnvp} extends NICE to non volume preserving layers that also perform a scaling operation. GLOW~\cite{kingma2018glow} further propose applying an invertible generalized permutation and a special normalization layer~(ActNorm) to improve expressivity.
\looseness=-1

\paragraph{Probabilistic models in RL.} Strides in probabilistic modeling research consistently push the boundaries in deep RL. Variational autoencoders gave rise to a family of model-based~\cite{hafner2019planet, hafner2024dreamerv3, kaiser2024simple} and representation learning~\cite{lee2020slac, Yarats_isac} algorithms. Contrastive learning has been used to learn value functions~\cite{ma2023vipuniversalvisualreward, sermanet2018tmw, eysenbach2023crl, myers2025tdc} as well as representations~\cite{srinivas2020curl, pmlr-v139-stooke21a} for RL. Advances in sequence modeling led to improvements in BC~\cite{chen2021dt, brohan2023rt1, lee2024vqbet, shafiullah2022cloningk}, RL~\cite{ni2022rmfrl, grigsby2024amago} and planning~\cite{janner2021dd}. More recently, diffusion models~\cite{song2021diff, ho2020diff, sohldickstein2015diff} and their variants~\cite{song2023consistencymodels, lipman2023fm, bengio2021gflow, bengio2023gflow} have become the popular models for large scale BC pre-training~\cite{chi2024diffpol, lin2025datascalinglawsimitation, pmlr-v164-weng22a, black2024pi0}, RL finetuning~\cite{ren2024dppo, hansenestruch2023idql}, offline RL~\cite{ajay2023cmayn, park2025fql, venkatraman2023ldcq} and planning~\cite{janner2022planningdiff, zheng2023guidedflows, lu2023synther}.
\looseness=-1

\paragraph{NFs in RL.} In RL, NFs have been used to learn policies from uncurated data~\cite{singh2020parrot} and model exact MaxEnt policies~\cite{chao2024meow, haarnoja2018lsp, chao2024meow}. But despite their flexibility~(see~\cref{fig:nf-capabilities} and~\cref{tab:nf-capabilities}), they have not been go-to models in RL; research exploring the application of NFs in RL remains limited. We believe this is because historically, NFs were thought to have limited expressivity~\cite{zhai2024tarf, papamakarios2021nfs}. In our experiments, we show that this is not the case~(\cref{sec:experiments}). We hope that our work brings greater attention towards the potential of NFs in RL, and the development of scalable probabilistic models with similar desirable properties~(\cref{fig:nf-capabilities}).
\looseness=-1

\section{Preliminaries}
This section describes two major approaches for training probabilistic models and how NFs can be trained using both of them. In the next section, we show that this flexibility allows NFs to be directly plugged into various RL algorithms, improving their expressivity and performance.

\label{sec:prelimns}
\paragraph{Maximum likelihood estimation (MLE).} Let $x \in \mathbb{R}^d$ be a random variable with an unknown density $p(x)$. Given a dataset of samples from this density $\mathcal{D} = \{x_i\}_{i=1}^{N} \sim p(x)$, the goal of MLE is to estimate the parameters of a probabilistic model~$p_{\theta}(x)$ that maximize the likelihood of the dataset:
\begin{equation}
\label{eq-mle}
   {\color{myblue} \theta_{\text{MLE}} = \argmax_{\theta} \E_{x \sim \mathcal{D}} \big[ \log p_\theta(x) \big] }.
\end{equation}
If $p_{\theta}(x)$ belongs to a universal family of distributions, then the MLE estimate $p_{\theta_{\text{MLE}}}(x)$ converges towards the true density, as dataset size~($N$) increases~\cite{Jordan2001AnIT}. 

\paragraph{Variational inference (VI).} In many problems, instead of samples from the true density $p(x)$, one has access to its unnormalized part $\tilde{p}(x) = p(x) \times Z$. The goal of variational inference is to approximate $p(x)$ using a distribution $q$ from some variational family of distributions $\gQ$:
\begin{equation}
\label{eq-vi}
    {\color{mypurple} q^*(x) = \argmin_{q \in \mathcal{Q}} D_\text{KL}(q \mid\mid p) = \argmax_{q \in \mathcal{Q}} \E_{x \sim q(x)} \big[ \log \tilde{p}(x) - \log q(x) \big]},
\end{equation}
where Z can be ignored from the last equation. If $\mathcal{Q}$ is universal, the global optima of Eq.~\ref{eq-vi} will be the true density $q^*(x) = p(x)$. To optimize Eq.~\ref{eq-vi} using gradient descent, one needs to choose a family $\mathcal{Q}$, from which it is easy generate samples~(the expectation requires sampling from $q$), evaluate likelihoods~(second term), and backpropagate through both these operations. Due to these constraints, one typically has to restrict $\mathcal{Q}$ to be a simpler family~\cite{Jordan1999vi}~(e.g., exponential family), or resort to other inference algorithms~\cite{graikos2023diffpnp} that require additional complexity~(more hyper-parameters or models). 
\looseness=-1

\paragraph{Normalizing flows (NFs).} NFs learn an invertible mapping $f_\theta : \mathbb{R}^d \rightarrow \mathbb{R}^d$, from true density $p(x)$ to the a simple prior density $p_0(z)$. This function is parameterized such that the inverse $f_{\theta}^{-1}$ as well as the determinant of its jacobian matrix~($| \frac{df_{\theta}(x)}{ dx}|$) is easily calculable. The learned density $p_{\theta}(x)$ can then be estimated using the change of variable formula:
\begin{equation}
\label{eq:nf-likelihood}
p_{\theta}(x) = p_0(f_{\theta}(x)) | \frac{df_{\theta}(x)}{ dx}|.    
\end{equation}
To sample from an NF, one first samples from the prior, and return the inverse function:
\begin{equation}
\label{eq:nf-sampling}
z \sim p_0(z), \; \text{and return} \; x = f_{\theta}^{-1}(z).
\end{equation}
Because \cref{eq:nf-likelihood} and \cref{eq:nf-sampling} are passes through a single neural network, back-propagating gradients through them is straightforward. Hence, the NF can be trained efficiently using both MLE~(\cref{eq-mle}) and VI~(\cref{eq-vi})~\cite{tabak2010nf, rezende2016nf}:
\begin{align}
   {\color{myblue} \theta_{\text{MLE}} } & {\color{myblue} = \argmax_{\theta} \E_{x \sim \mathcal{D}} \big[ \log p_0(f_{\theta}(x)) + \log ( | \frac{df_{\theta}(x)}{ dx}| ) \big] }. \label{eq-nf-mle} \\
   {\color{mypurple} \theta_{\text{VI}}} & {\color{mypurple} = \argmax_{\theta} \E_{ \substack{ z \sim p_0(z) \\ x = f^{-1}_{\theta}(z) } } \big[ \log \tilde{p}(x) - \log p_0(z) + \log ( | \frac{df^{-1}_{\theta}(x)}{ dx}| ) \big] }. \label{eq-nf-vi}
\end{align}
\Cref{eq-nf-mle} follows from substituting the change of variable formula in \cref{eq-mle}. \Cref{eq-nf-vi} is derived by the same substitution, followed by noting that $z = f_{\theta}( f^{-1}_{\theta}(z) )$ and $ - \log ( |\frac{df^{-1}_{\theta}(x)}{ dx}|) = \log ( |\frac{df_{\theta}(x)}{ dx}| )$. It is standard practice to stack multiple such mappings together $f_{\theta} = f_{\theta}^1 \circ f_{\theta}^1 \circ \dots \circ f_{\theta}^T$. Each mapping is referred to as a \textit{block} in the entire NF.
\looseness=-1

\section{Many RL Algorithms do MLE and VI.}
\label{sec:rl-vi-mle}
In this section, we first introduce multiple RL problems settings. We will then show how popular representative algorithms in each setting are instantiations of MLE and VI. This will $(1)$~underscore the importance of flexible models that can be directly optimized using MLE and VI, and $(2)$~motivate our use of NFs as a generally capable family of models for RL. We will then describe our architecture in the next section~(\cref{sec:nf-arch}). In all the following problems, we assume an underlying controlled markov decision process~(MDP, $\mathcal{M}$), with states $s \in \mathcal{S}$ and actions $a \in \mathcal{A}$. The dynamics are $p(s^\prime | s, a)$, the initial state distribution is $p_0(s_0)$ and $H$ is the episode horizon.
\looseness=-1

\subsection{Offline Imitation Learning (IL)}
Given a dataset of state-action trajectories $\mathcal{D} = \{s_0^i,a_0^i, \dots, s_H^i\ \}_{i=1}^N$ collected by an expert(s), the aim in IL is to match the data generating behavior without any environment interactions.

\paragraph{Behavior Cloning (BC)~\cite{Bain1995_bc, Pomerleau_bc}}\phantomsection\label{alg:bc} is an algorithm that models the conditional distribution of actions given states directly using MLE:
\begin{equation}
\label{eq-bc}
   \argmax_{\theta} {\color{myblue} \underbrace{\E_{s_t,a_t \sim \mathcal{D}} \big[ \log {\pi}_\theta(a_t \mid s_t) \big]}_{ \text{MLE} } }.
\end{equation}
\paragraph{Goal Conditioned Behavior Cloning (GCBC)~\cite{emmons2022rvs, ding2020gcbc, ghosh2020gcbc}}\phantomsection\label{alg:gcbc}is a simple variant of BC which uses a goal $g \in \mathcal{S}$ as additional conditioning information. It is based on the insight that a trajectory leading to any goal, can be treated as an optimal trajectory to reach that goal. Given a state-action pair, a state sampled from the future~$p_{t+}(s_{t+} \mid s_t,a_t)$ is treated as the goal. Formally, GCBC maximizes the following objective:
\begin{equation}
\label{eq-gcbc}
   \argmax_{\theta} {\color{myblue} \underbrace{ \E_{ \substack{s_t,a_t \sim \mathcal{D} \\ g_{t+} \sim p(s_{t+} \mid s_t,a_t) } } \big[ \log {\pi}_\theta(a_t \mid s_t, g_{t+}) \big] }_{ \text{MLE} } }.
\end{equation}
Following prior work~\cite{park2025ogbench, eysenbach2023ocbc}, we set the future time $t+$ to be a truncated geometric distribution (from $t+1$ to $H$).
\looseness=-1

\subsection{Offline RL}
Offline RL assumes an MDP~$\mathcal{M}$ with a reward function~$r(s, a)$. Given a dataset of trajectories $\mathcal{D} = \{s_0^i,a_0^i, r^i_0, \dots, s_H^i, r_H^i\ \}_{i=1}^N$, the goal is to find a policy $\pi_{\theta}(a \mid s)$, which maximizes the expected discounted return $\E_{\pi_{\theta}, p} [ \sum_{t=0}^{H} \gamma^t r(s_t, a_t)]$~\cite{levine2020orl, kumar2019orl}.

\paragraph{MaxEnt RL + BC~\cite{fujimoto2021minorl, levine2018cic, haarnoja2018sac}.}\phantomsection\label{alg:maxent-rl} We write out a minimal objective for an offline actor-critic algorithm~\cite{fujimoto2021minorl}:
\begin{align}
\label{eq:maxent-rl}
 	&\argmin_{Q_\phi} \E_{\substack{s_t,a_t,r_t,s_{t+1} \sim D \\ a_{t+1} \sim \pi_{\theta}(\cdot \mid s_{t+1})}} \left[\left( Q_{\phi}(s_t,a_t) - r_t - \gamma Q_{\overbar{\phi}}(s_{t+1},a_{t+1}) \right)^2 \right],\\
 	&\argmax_{ \pi_\theta } {\color{mypurple} \underbrace{ \E_{s_t \sim D, a^{\pi}_t \sim \pi_{\theta}(\cdot \mid s_t)} \big[ Q_{\phi}(s_t,a^{\pi}_t) - \lambda \log \pi_{\theta}(a^{\pi}_t \mid s_t) \big] }_{ \text{VI} } } +  {\color{myblue} \underbrace{ \alpha \E_{s_t, a_t \sim D} \big[ \log \pi_{\theta}(a_t \mid s_t) \big] }_{ \text{MLE} } },
\end{align}
where $Q_\phi(s,a)$ is the action-value function and $Q_{\overbar{\phi}}(s,a)$ denotes the target network~\cite{mnih2013dqn}, $\lambda$ and $\alpha$ control the strength of the entropy and behavior regularization respectively. For $\lambda = 0$, this algorithm exactly matches TD3+BC~\cite{fujimoto2021minorl}. In all our experiments, we use $\lambda = 0$ for minimality\footnote{This is a special case of variational inference. In bayesian inference this corresponds to approximating a maximum a-posteriori estimate.}.
The policy has to sample actions that have high values while staying close to the (potentially multi-modal) behavior policy. 
NFs ability to be trained using both MLE and VI allows the use of expressive policies, without \textit{any changes} to the simplest offline RL algorithms. 

\looseness=-1
\subsection{Goal conditioned RL (GCRL)}
GCRL~\cite{Kaelbling1993LearningTA} is a multi-task RL problem where tasks correspond to reaching goals states $g \in \mathcal{S}$ sampled from a test goal distribution $p_{\text{test}}(g)$. We define the discounted state occupancy distribution as $p^{\pi}_{t+}(s_{t+} \mid s,a) = (1-\gamma) \sum_{t=0}^{\infty} \gamma^t p(s_t = s_{t+} \mid s_0,a_0=s,a)$.\footnote{We will work with in infinite horizon case, with $H=\infty$.}

\paragraph{Q-function estimation.} Prior work has shown that the Q-function for GCRL is equal to the discounted state occupancy distribution $Q^{\pi}(s,a,g) = p^{\pi}_{t+}(s_{t+} \mid s,a)$~\cite{eysenbach2023crl}. Hence estimating the Q-function can be framed as a maximum likelihood problem:
\begin{equation}
\label{eq-gcve}
   \argmax_{\theta} {\color{myblue} \underbrace{ \E_{ \substack{s_t,a_t \sim \mathcal{D} \\ g_{t+} \sim p_{t+}(s_{t+} \mid s_t,a_t) } } \big[ \log p_\theta( g_{t+} \mid s_t, a_t) \big] }_{ \text{MLE} } }.
\end{equation}

\paragraph{Unsupervised goal sampling (UGS)~\cite{florensa2018ag, pitis2020goalkde}}\phantomsection\label{alg:ugs}  algorithms do not assume that the test time goal distribution $p_{\text{test}}(g)$ is known a priori. The main aim of UGS is to maximize exploration by commanding goals that are neither too easy nor too difficult for the current agent. This inclines the agent to explore goals on the edge of its goal coverage. But to recognize which goals are on the edge, one needs an estimate of the goal coverage density $p_{t+}(g)$. Many methods~\cite{pong2020skewfits, pitis2020goalkde, florensa2018ag} primarily fit a density model and then choose goals that have low density. A natural way of estimating densities is MLE, though some prior works use secondary objectives~\cite{openai2021asym, sukhbaatar2018imself, florensa2018ag, pong2020skewfits} or non-parametric models~\cite{pitis2020goalkde}:
\begin{equation}
\label{eq-gccde}
   \argmax_{\theta} {\color{myblue} \underbrace{ \E_{ \substack{s_t,a_t \sim \mathcal{D} \\ g_{t+} \sim p_{t+}(s_{t+} \mid s_t,a_t) } } \big[ \log p_\theta(g_{t+}) \big] }_{ \text{MLE} } }.
\end{equation}
Using an estimate of this density, the canonical UGS algorithm is to sample a large batch of goals from the replay buffer and command the agent to reach the goal with the minimum density under~$p_\theta(g_{t+})$~\cite{pitis2020goalkde}.

\looseness=-1
\section{A Simple and Efficient NF Architecture for RL}
\label{sec:nf-arch}
This section will introduce the NF architecture we use and~\cref{sec:experiments} will show how the this architecture fits seamlessly into many RL algorithms (those from~\cref{sec:rl-vi-mle}). Many design choices affect model expressivity, the speed of computing the forward~($f_\theta(x)$) and inverse~($f_\theta^{-1}(x)$), and the efficiency of computing the determinant of the Jacobian~\cite{papamakarios2021nfs}.
Our desiderata is to build a \textit{simple} and \textit{fast} architecture that is \textit{expressive enough} to compete and even surpass the most popular approaches across a range of RL problems. To achieve this, we build each NF block $f^t_{\theta}$ by leveraging two components from prior work: \textit{(1)} a fully connected ``coupling'' network~\cite{dinh2017realnvp}, and \textit{(2)} a linear flow~\cite{kingma2018glow}.
% In~\cref{fig:nf-block}, we visualize this NF block.
In all of our experiments, the model learns a conditional distribution $p_{\theta}(x \mid y)$, so each NF block $f^t_{\theta} :  \mathbb{R}^d \times \mathbb{R}^k \rightarrow \mathbb{R}^d$ takes in $y \in \mathbb{R}^k$ as an additional input.

\paragraph{Coupling network.} We use an expressive and simple transformation proposed in~\citet{dinh2015nice, dinh2017realnvp}. Let $x_t$ be the input for the $t^\text{th}$ block and $y$ be the conditioning information. For the $t^\text{th}$ block, the coupling network $f_{\theta}^t = \{ a_{\theta}^t, s_{\theta}^t : \mathbb{R}^{\lfloor d/2 \rfloor} \times \mathbb{R}^k \rightarrow \mathbb{R}^{\lceil d/2 \rceil} \}$ does the following computation:
\begin{align*}
	&(1)\; x_1^t, x_2^t = \text{split}(x^t),\;(2)\;\tilde{x}_2^{t} = ( x_2^t + a_{\theta}^t(x_1^t, y)) \times \exp(- s_{\theta}^t(x_1^t, y) ),\;(3)\;\tilde{x}^{t} = \text{concat}(x_1^t, \tilde{x}_2^t).
\end{align*}
Note that the log-det Jacobian of $f^t_{\theta}$ is $-\text{sum}(s_{\theta}^t(x_1^t, y))$. Further, given $\tilde{x}^{t}$ and $y$, the inverse ${f^{t}_{\theta}}^{-1}$ can be calculated as easily as the forward map (see~\cref{app:nf-details} for details).

\looseness=-1
\paragraph{Linear flow.} We adapt the generalized permutation proposed by~\citet{kingma2018glow} to one dimensional inputs, to linearly transform the output of the coupling layer $\tilde{x}^{t}$. The intuition behind adding this linear flow is that each dimension of $\tilde{x}^t$ can be transformed by other dimensions in an end to end manner:
\begin{align*}
	&x^{t+1} = (\tilde{x}^{t})^T W,\; \text{where} \; W=PL_{\theta}U_{\theta}, \; \text{and} \; \log|W| = \text{sum} \log (|\text{diag}(U)|).
\end{align*}

Here, $P$ is a fixed permutation matrix and $L_{\theta}$ is lower triangular with ones on the diagonal and $U_{\theta}$ is upper triangular. Hence, inverting them incurs similar costs as matrix multiplication~\cite{papamakarios2021nfs} and calculating the log-det Jacobian requires only the diagonal elements of $U$. 
\looseness=-1

\paragraph{Normalization.} While previous approaches used special variants of normalization layers~(\cite{dinh2017realnvp}, \cite{kingma2018glow}), we find that just using LayerNorm~\cite{ba2016ln} in the coupling network's weights was sufficient of training deep NFs.

Overall, our architecture can be thought of as RealNVP~\cite{dinh2017realnvp} with generalized permutation, or a linear version of GLOW~\cite{kingma2018glow} without ActNorm. We have released a simple code for this architecture both in Jax and PyTorch.
\looseness=-1

\section{Experiments}
\label{sec:experiments}

\begin{wrapfigure}{r}{0.5\textwidth} 
    \centering
    \vspace{-1.5em}
    \includegraphics[width=\linewidth]{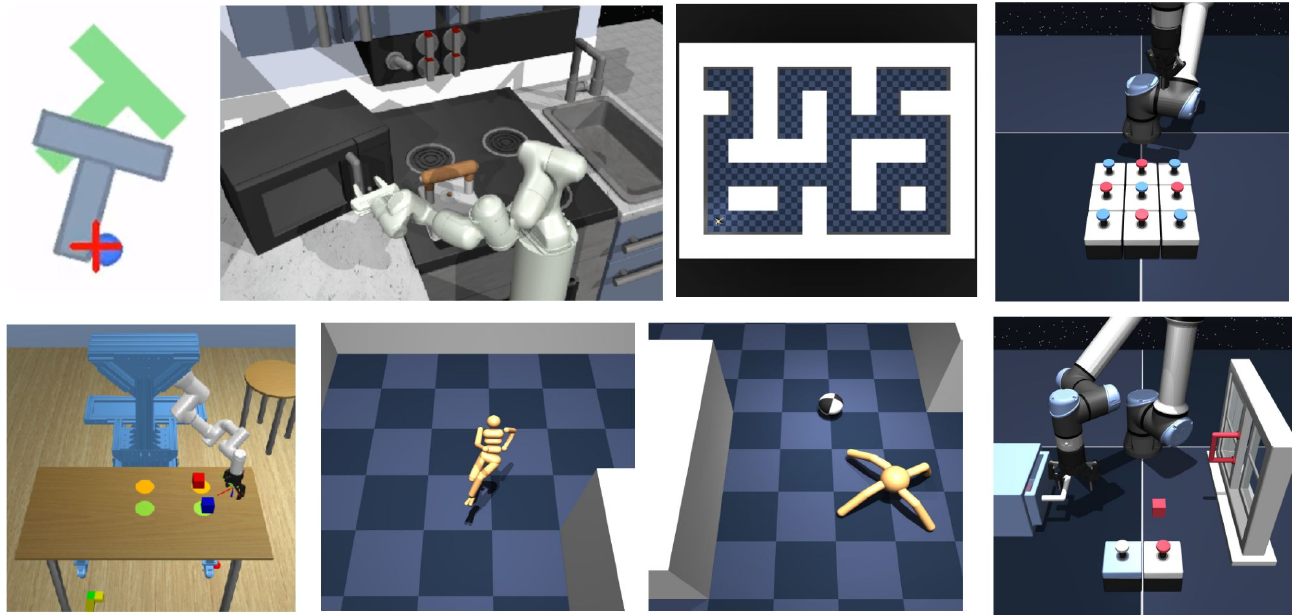}
    \vspace{-1.5em}
    \caption{\footnotesize \textbf{Visualization of tasks}}
    \label{fig:task-viz}
    \vspace{-1em}
\end{wrapfigure}

The main aim of our experiments is to use NFs and question some of the standard assumptions across several RL settings. $(1)$ In IL, methods like diffusion policies~\cite{chi2024diffpol} are popular because they provide expressivity but at a high computational cost. Our experiments with NFs show that this extra compute might not be necessary~(\cref{sec:experiments-il}). $(2)$ In conditional IL, prior work argues that learning a value function is necessary for strong performance~\cite{kumar2020cql, chebotar2023qtransf, fujimoto2019bcq, agarwal2020opt}. Our experiments suggest that a sufficiently expressive policy, such as an NF, can often suffice~(\cref{sec:experiments-cil}). $(3)$ In offline RL, auxiliary techniques are typically used to support expressive policies. We show that NF policies can be directly integrated with existing offline RL algorithms to achieve high performance~(\cref{sec:experiments-orl}). $(4)$ In UGS, we show that a single NF can estimate both the value function and goal density, enabling better exploration compared to baseline that make stronger assumptions. 

In all experiments, we use the architecture described in \cref{sec:nf-arch} to implement all algorithms described in \cref{sec:rl-vi-mle}. Implementation details for all algorithms are provided in \cref{app:implementation}. \textit{Across all problems~(IL, offline RL, online GCRL), we present results on a total of {82} tasks~(see~\cref{fig:task-viz} for visualization of tasks) with 5 seeds each, which show that NFs are generally capable models that can be effectively applied to a diverse set of RL problems.} Unless stated otherwise, all error bars represent $\pm$ standard deviation across seeds. Finally, in~\cref{app:ablation-results}, we provide ablation experiments for various design choices and hyper-paramter values for some of the algorithms presented in this paper.

\subsection{Imitation Learning (IL)}
\label{sec:experiments-il}

\begin{figure}[h]
\centering 
\begin{subfigure}[b]{\textwidth}
    \centering
    \includegraphics[width=\linewidth]{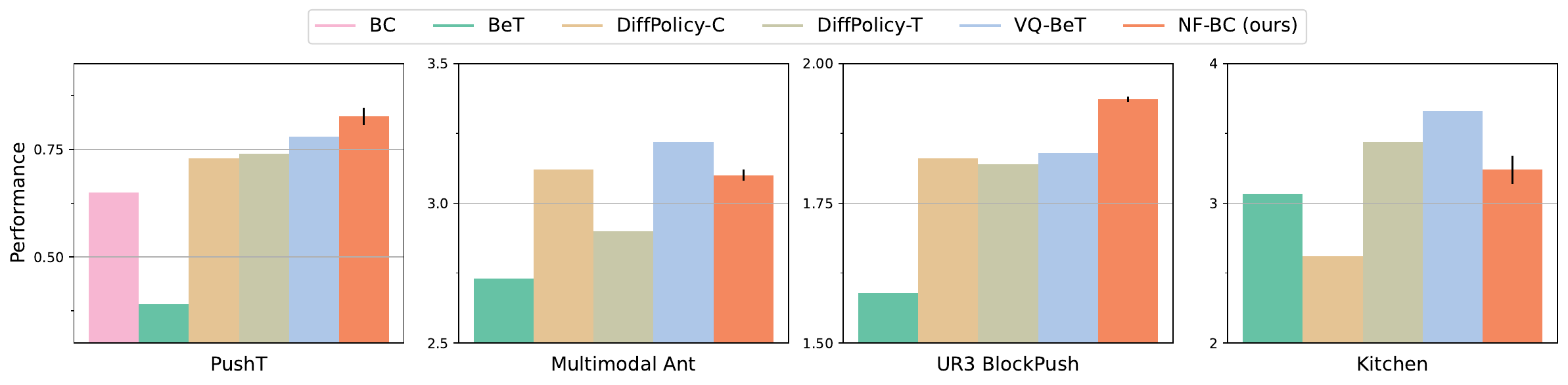} 
    \caption{ \footnotesize \textbf{Performance comparison.} Performance is measured using normalized rewards.}
    \vspace{10pt}
    \label{fig:il-result}
\end{subfigure}
\begin{subfigure}[b]{\textwidth}
    \centering
    \begin{tabular}{cccccc}
        \toprule
        \textbf{\makecell[t]{Method}} & \textbf{\makecell[t]{Time (sec) \\ inference; training}} & \textbf{\makecell[t]{Params (M)}} & \textbf{\makecell[t]{Model types}} & \textbf{\makecell[t]{Objectives}} \\
        \midrule
        NF-GCBC          & 0.127 ; 0.023 & 10 M      & Fully connected NFs & 1; end to end \\
        VQ-BET           & 0.273 ; 0.021 & 4.23 M    & Transformer + VQ-VAEs & 2; req. pre-train \\
        Diffusion Policy & 1.28  ; 0.019 & 65.78 M   & Convolutional network & 1; end to end \\
        \bottomrule
    \end{tabular}
    \caption{\footnotesize \textbf{Conceptual comparison} to provide a more holistic view for RL practitioners.}
    \vspace{10pt}
    \label{tab:il-result}
\end{subfigure}
\caption{\footnotesize \textbf{Imitation Learning.} Results on multi-modal behavior datasets which require fine
grained control and expressivity. NF-BC is faster in inference, easier to implement (end to end optimization of a single loss function, contains 2-3 times less hyper-parameters ) and achieves competitive performance across all tasks (best success rate on two tasks and in top 2 methods on the other two tasks). In~\cref{fig:il-result}, we only report standard errors for our method as baseline results were taken from prior work which did not include error bars~\cite{lee2024vqbet}}
\end{figure}

In this section, we use our NF architecture with BC~(\cref{alg:bc}, \cref{eq-bc}) for imitation learning~(NF-BC). Please see implementation details of training and sampling with IL algorithm in~\cref{app:implementation-il}.

\textbf{Tasks.} PushT requires fine-grain control to maneuver a T-block in a fixed location. Performance is measured by the final normalized area of intersection between the block and the goal. Multimodal Ant, UR3 BlockPush, and Kitchen contain multiple behaviors~(4, 2, and 4 respectively). Performance is measured by the normalized number of distinct behaviors the agent is able to imitate successfully over 50 independent trials. We also report the simplicity of algorithms, as it is often a decisive factor in determining whether the algorithm can be practically employed. Simplicity is measured by the negating the number of hyperparameters used. We list and briefly describe all the hyperparameters of all the methods used to make \cref{fig:il-result} in ~\cref{app:hps-baseline}. \textbf{Baselines.} We compare our algorithms with 4 baselines. Diffusion policies~\cite{chi2024diffpol} use high performing diffusion model architectures. We use both the convolutional and transformer~(DiffPolicy-C and DiffPolicy-T) variant of the diffusion policy. VQ-BeT~\cite{lee2024vqbet} uses VQ-VAEs~\cite{oord2018vqvae} for discretizing actions and learns a transformer model over chunked representations. BeT~\cite{shafiullah2022cloningk} is similar to VQ-BeT but uses K-means for action chunking.

\textbf{Results.} In~\cref{fig:il-result} we see that NF-BC outperforms all baselines on PushT and Kitchen and performs competitively in other two tasks.  Of all the baselines, NF-BC is fastest in inference and easier to implement (\cref{tab:il-result}). For example, DiffPolicy-T uses 3 times more hyperparameters than NF-BC. These results suggest that NF policies do not lack in expressivity, compared to both diffusion and auto-regressive policies despite introducing fewer additional hyper-parameters. In fact, NF-BC can achieve these results using just a fully connected network and fraction of parameters compared to diffusion policies (see rightmost plot in \cref{fig:il-result}). Unlike BeT~\cite{shafiullah2022cloningk} and VQ-BeT~\cite{lee2024vqbet}, NF-BC does not use autoregressive models or VQ-VAEs (see~\cref{tab:il-result} for a comparison of architectures). Reducing the number of hyperparameters is especially important for practitioners because it makes implementation easier and decreases the need for hyperparameter tuning.

\subsection{Conditional Imitation Learning}
\label{sec:experiments-cil}
In this section, we use our NF architecture with GCBC~(\cref{alg:gcbc}, \cref{eq-gcbc}) for conditional imitation learning~(NF-GCBC).

\textbf{Tasks.} We use a total of 45 tasks from OGBench~\cite{park2025ogbench} meant to test diverse capabilities like learning from suboptimal trajectories with high dimensional actions, trajectory stitching and long-horizon reasoning. Performance is measured using the average success rate of reaching a goal in 50 independent trials.
\textbf{Baselines.} We compare with 6 different baselines. GCBC is the GCBC~\ref{alg:gcbc} algorithm with a gaussian policy, and FM-GCBC is the GCBC~\ref{alg:gcbc} algorithm with a flow-matching based policy~\cite{lipman2023fm}. The velocity field for FM-GCBC is a fully connected network trained using flow matching. FM-GCBC is the closest baseline to NF-GCBC as it uses the same underlying algorithm, uses a fully connected network like NF-GCBC, and uses an expressive family of policy models. GCIQL~\cite{kostrikov2021iql} is an offline RL algorithm that learns the optimal value function corresponding to the best actions in the dataset. QRL~\cite{wang2023qrl} uses a quasimetric architecture to learn the optimal value function and CRL~\cite{eysenbach2023crl} uses contrastive learning to estimate behavior policy's value function.

\begin{figure}[h]
	\centering
	\includegraphics[width=\linewidth]{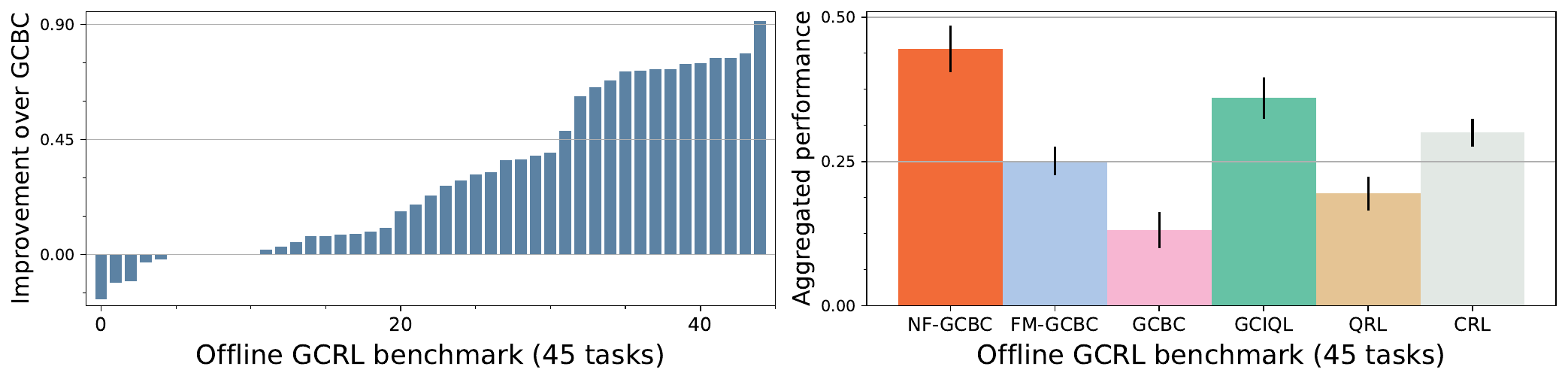}
	\vspace{-0.5em}
	\caption{\label{fig:gcil-result} \footnotesize \textbf{Conditional Imitation Learning.} \figleft \, NF-GCBC exceeds its gaussian counterpart GCBC on ${40}/{45}$ tasks, demonstrates that using expressive models like NFs can significantly improve performance. \figright \, Despite being a BC–based algorithm, NF-GCBC outperforms all offline RL baselines across 45 tasks. NF-GCBC performs 77\% better than FM-GCBC which is the closest baseline that uses a flow matching policy with GCBC.}
	\vspace{-1em}
\end{figure}

\textbf{Results.} In \cref{fig:gcil-result} (left), we see that NF-GCBC exceeds the performance (often significantly) of GCBC on ${40}/{45}$ tasks. This underscores the importance of using an expressive policy with the GCBC algorithm. In \cref{fig:gcil-result} (right), NF-GCBC, despite being a BC based algorithm, outperforms all baselines, including dedicated offline RL algorithms. We have added separate task wise plots in \cref{app:additional-results}. NF-GCBC outperforms FM-GCBC, which is the closest baseline, by 77\%. While this does not necessarily imply that NFs are more expressive than flow-matching models, we hypothesize that all else being equal~(e.g., fully connected architectures and identical underlying algorithms), NFs are better able to imitate the data using simple architectures as they directly optimize for log likelihoods. Flow matching models on the other hand estimate probabilities indirectly (by learning an SDE~/~ODE), and might require more complex architectures~(transformers or convolutional networks) to work well.

\subsection{Offline RL}
\label{sec:experiments-orl}

\begin{figure}[h]
	\centering
	\vspace{-1em}
	\includegraphics[width=\linewidth]{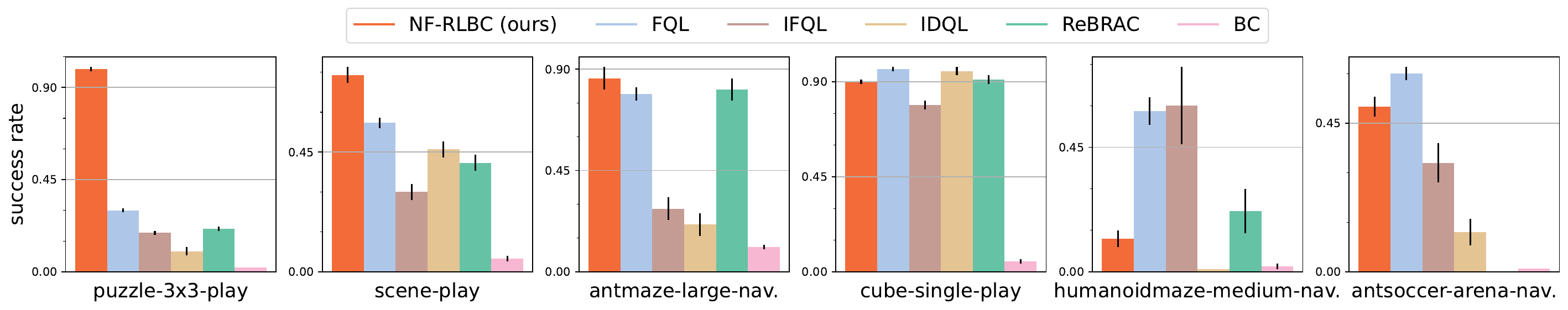}
    \vspace{-1em}
	\caption{\label{fig:offlinerl-result} 
    \footnotesize \textbf{Offline RL.} NF-RLBC outperforms all baselines on ${15}/{30}$ tasks, and achieves top 2 results on ${10}/{30}$ tasks. Notably, NF-RLBC achieves much higher results than all baselines on puzzle-3x3-play and scene-play, which require the most long horizon and sequential reasoning. For example, on puzzle-3x3-play NF-RLBC performs 230\% better than the next best baseline.}
	% \vspace{-1em}
\end{figure}

While NF-GCBC performs better than offline RL algorithms on average, there are certain tasks like puzzle-3x3-play where only TD based methods succeed~(\cref{app:additional-results}). This task, and many others as we will see, require learning a value function and accurately extracting behaviors from it. Fortunately, NFs can be trained using both VI and MLE~(\ref{sec:prelimns}). This allows us to train MaxEnt RL+BC~(\cref{alg:maxent-rl}, \cref{eq:maxent-rl}) using our NF architecture as a policy (NF-RLBC).

\textbf{Tasks.} We use a total of 30 tasks from prior work~\cite{park2025fql} testing offline RL with expressive policies. Each task has a goal which requires completing $K$ subtasks. At each step, the agent gets a reward equal to the negative of the number of subtasks left to be completed. For our experiments, we choose tasks that cover a diverse set of challenges and robots. Tasks like humanoidmaze-medium-navigate and antmaze-medium-navigate contain diverse suboptimal trajectories and high dimensional actions. Antsoccer-arena-navigate requires controlling a quadruped agent to dribble a ball to a goal location. Scene-play requires manipulating multiple objects and puzzle-3x3-play requires combinatorial generalization and long-horizon reasoning. We hypothesize that the Q function for the some of these tasks~(esp manipulation tasks like puzzle and scene which require long horizon reasoning) can have narrow modes of good actions and a large number of bad actions. Hence using direct gradient based optimization can be crucial to search for good actions. \textbf{Baselines.} We choose 5 representative baselines. BC performs BC with a gaussian policy, ReBRAC~\cite{rebrac} is an offline RL algorithm with a gaussian policy that achieves impressive results on prior benchmarks~\cite{fu2021d4rl}. IDQL~\cite{hansenestruch2023idql} uses importance resampling to improve over a diffusion policy. IFQL is
the flow counterpart of IDQL~\cite{park2025fql}. FQL~\cite{park2025fql} is a strong offline RL algorithm that distills a flow-matching policy into a one-step policy, while jointly maximizing the Q-function. Notably, all offline RL algorithms (IDQL, IFQL, FQL) have to incorporate additional components—such as distillation or resampling—to leverage policy expressiveness.

\textbf{Results.} \Cref{fig:offlinerl-result}, shows that on average NF-RLBC outperforms all baselines on ${15}/{30}$ tasks, and achieves top 2 results on ${10}/{30}$ tasks. Notably, NF-RLBC achieves much higher results than all baselines on puzzle-3x3-play and scene-play, which require most long horizon and sequential reasoning (on puzzle-3x3-play NF-RLBC performs 230\% better than the next best baseline). NF-RLBC is able to search for good actions because it directly uses gradient based optimization to backpropagate action gradients through the policy. All other offline RL algorithms with expressive policy classes~(IDQL, IFQL, FQL) have to use extra intermediate steps.

\begin{figure}[t]
	\centering
	\vspace{-1em}
	\includegraphics[width=\linewidth]{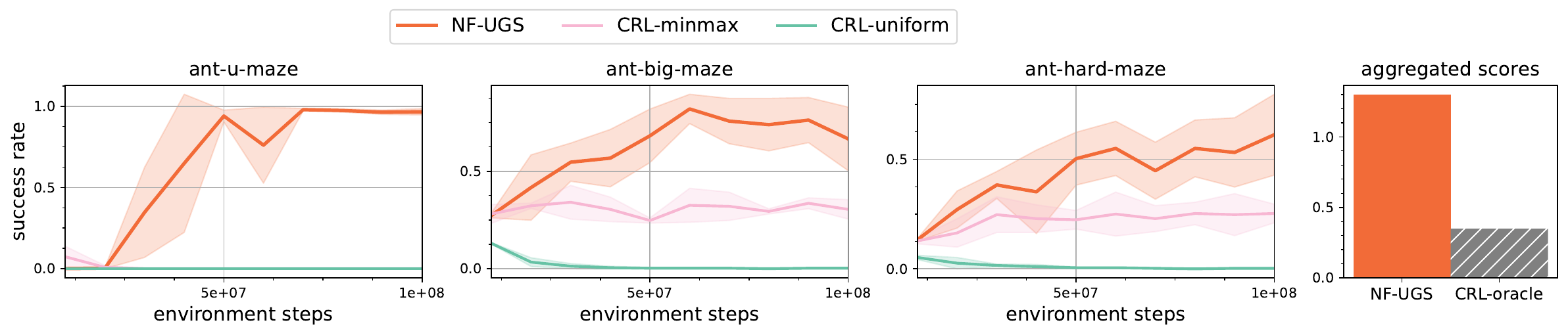}
    \vspace{-1em}
	\caption{\label{fig:gcrl-result} \footnotesize \textbf{GCRL.} (first 3 from left) NF-UGS consistently outperforms other unsupervised baselines. (rightmost) NF-UGS achieves higher asymptotic success rates than the CRL-oracle averaged across all tasks. Note that CRL-oracle makes extra assumptions about the availability of test-time goals.}
	\vspace{-1em}
\end{figure}

\subsection{Goal Conditioned RL}
\label{sec:experiments-gcrl}
In this section, we use our NF architecture to estimate both the Q-function~(\cref{eq-gcve}) and the coverage density~(\cref{eq-gccde}). Both quantities model the distribution over the same random variable~(future goals), with the Q-function conditioned on a state-action pair. We use a single neural network to estimate both by employing a masking scheme to indicate the presence or absence of conditioning~\cite{ho2022cfg}. We then train a goal-conditioned policy to maximize the Q-function in a standard actor-critic setup, and sample training goals using the canonical UGS algorithm~(\cref{alg:ugs}), resulting in our NF-UGS method.

\textbf{Tasks.} We use three problems from the UGS literature~\cite{pitis2020goalkde, bortkiewicz2024jaxgcrl} ant-u-maze, ant-big-maze and ant-hard-maze. These environments are difficult because the goal test distribution is unknown, and success requires thorough exploration of the entire maze. \textbf{Baselines.} We compare against three baselines. CRL-oracle~\cite{eysenbach2023crl, bortkiewicz2024jaxgcrl} uses contrastive learning to estimate the Q-function and serves as an oracle baseline as it uses test goals $p_\text{test}(g)$ to collect data during training. CRL-uniform is a variant that samples goals uniformly from the replay buffer, while CRL-minmax selects goals with the lowest Q-value from the initial state of the MDP -- intuitively targeting the goals the agent believes are hardest to reach. The aim of these comparisons isn't to propose a state of the art method, but rather to check whether NFs can efficiently estimate the coverage densities of the non-stationary RL data for unsupervised exploration. 
\textbf{Results.} These experiments aim to evaluate whether NFs can efficiently estimate both the Q-function and the coverage density. As shown in~\cref{fig:gcrl-result}, NF-UGS outperforms all baselines, and notably the oracle algorithm as well~(aggregated score is the average of the final success rates across all three tasks). Unlike the minmax strategy, which plateaus over time, NF-UGS continues to improve by sampling goals to maximize the entropy of the coverage distribution~\cite{pitis2020goalkde}.
\looseness=-1

\section{Conclusion}
\label{sec:conclusion}
In this paper, we view RL problems from a probabilistic perspective, asking: which probabilistic models provide the capabilities necessary for efficient RL? We argue that NFs offer a compelling answer. NFs outperform strong baselines across diverse RL settings, often with fewer parameters and lower compute, leading to simpler algorithms. By making density estimation simpler and effective, we hope our work opens new avenues for areas such as distributional RL, bayesian RL, and RL safety.

A primary limitation of NFs is that they have restrictive architectures, in that they have to be invertible. Nevertheless, there is a rich line of work characterizing which NF architectures are universal~\cite{draxler2025uni, cornish2021rel, papamakarios2021nfs}. Another limitation of our work is that we do not explore new NF architectures; the design we use is mainly adapted from prior work~(\cref{sec:nf-arch}).

\begin{ack}
The authors are pleased to acknowledge that the work reported on in this paper was substantially performed using the Princeton Research Computing resources at Princeton University which is consortium of groups led by the Princeton Institute for Computational Science and Engineering (PICSciE) and Office of Information Technology's Research Computing. The authors thank Emmanuel Bengio, Liv d'Aliberti and Eva Yi Xie for their detailed reviews on our initial drafts. The authors thank Seohong Park for providing code for some baselines and Catherine Ji for helping find a bug in our code. The authors also thank Shuangfei Zhai and the members of Princeton RL lab for helpful discussions.
\end{ack}

%%%%%%%%%%%%%%%%%%%%%%%%%%%%%%%%%%%%%%%%%%%%%%%%%%%%%%%%%%%%

\newpage

\appendix

\section{Implementation details.}
\label{app:implementation}
In this section we provide all the implementation details as well as hyper-parameters used for all the
algorithms in our experiments – NF-BC, NF-GCBC, NF-RLBC, and NF-UGS. Unless specified otherwise, we chose common hyper-parameters used in prior work without tuning. 

\subsection{Imitation Learning: NF-BC}
\label{app:implementation-il}
For all imitation learning, we use a policy with 12 NF blocks. The fully connected networks of the coupling network for each NF block consist of two hidden layers with 512 activations each. We use a fully connected network for the state-goal encoder with 4 hidden layers having 512 activations each.   

It is common in the NF literature to add some noise to the training inputs before training an NF using MLE~\cite{rezende2016nf, zhai2024tarf}. For NF-BC, we add gaussian noise to the actions before training. Hence, NF-BC models the noisy behavior policy instead~$\pi^\text{a+noise}(a \mid s)$. To reduce noise from the obtained samples, ~\citet{zhai2024tarf} propose a denoising trick:

\begin{equation*}
    z \sim p_0, \; y:=f^{-1}(z) \;, x:= y + \sigma^2 \nabla_y \log p_{\theta}(x) 
\end{equation*}

We use this trick in all our NF-BC experiments. \Cref{tab:hp-values-nfbc} contains the values of hyperparameters used by NF-BC. All experiments were done on RTX 3090s, A5000s or A6000s and did not require more than 20 hours. Experiment speed is mostly bottlenecked by intermediate evaluation of BC policies because the environments are slow and the number of rollouts per evaluation is 50. For example, the experiments on the multimodal-ant environment took 20 hours, but on pusht they were finished in less than 90 minutes.  

\begin{table}[h]
    \centering
    \caption{\footnotesize \textbf{NF-BC:} Hyper-parameters and their values.} 
    \begin{tabular}{l c}
        \toprule
        \text{Hyper-parameter} & \text{Description} \\
        \midrule
        channels & 512 \\
        noise std & 0.1 \\
        blocks & 12 \\
        encoder num layers & 4 \\
        rep dims & 512 \\
        \bottomrule
        \end{tabular}
    \label{tab:hp-values-nfbc}
\end{table}

\subsection{Conditional Imitation Learning: NF-GCBC}
\label{app:implementation-cond-il}
For all conditional imitation learning experiments, we use a policy with 6 NF blocks. The fully connected networks of the coupling network for each NF block consist of two hidden layers with 256 activations each. We use a fully connected network for the state-goal encoder with 4 hidden layers having 512 activations each. The output representation dimensions are 64. We also employ the denoising trick proposed by \cite{zhai2024tarf} and mentioned in~\cref{app:implementation-il}. \Cref{tab:hp-values-nfgcbc} contains the values of hyperparamters used by NF-GCBC. All experiments were done on RTX 3090s, A5000s or A6000s and did not require more than 3 hours each. 

\begin{table}[h]
    \centering
    \caption{\footnotesize \textbf{NF-GCBC:} Hyper-parameters and their values.} 
    \begin{tabular}{l c}
        \toprule
        \text{Hyper-parameter} & \text{Description} \\
        \midrule
        channels & 512 \\
        noise std & 0.1 \\
        blocks & 6 \\
        encoder num layers & 4 \\
        rep dims & 512 \\
        Future goal sampling discount & \Cref{tab:gamma-nfgcbc} \\
        \bottomrule
        \end{tabular}
    \label{tab:hp-values-nfgcbc}
\end{table}

\begin{table}[h]
    \centering
    \caption{\footnotesize \textbf{NF-GCBC:} Environment values for future goal sampling discount values. For all GCBC algorithms, including the GCBC baselines, we tuned this value in the set \{0.97, 0.99\}} 
    \begin{tabular}{l c}
        \toprule
        \text{Environment Name} & \text{Future goals sampling discount value} $\gamma$ \\
        \midrule
        puzzle $3\times3$ play & 0.97 \\
        scene play & 0.97 \\
        cube single play & 0.97 \\
        cube double play & 0.99 \\
        humanoidmaze medium navigation & 0.97 \\
        antmaze large navigation & 0.99 \\
        antsoccer arena navigation & 0.97 \\
        antmaze medium stitch & 0.97 \\
        antsoccer arena stitch & 0.97 \\
        \bottomrule
        \end{tabular}
    \label{tab:gamma-nfgcbc}
\end{table}

\subsection{Offline RL: NF-RLBC}
\label{app:implementation-offline-rl}
For all offline RL experiments, we use a policy with 6 NF blocks. The fully connected networks of the coupling network for each NF block consist of two hidden layers with 256 activations each. We use a fully connected network for the state-goal encoder with 4 hidden layers having 512 activations each. The output representation dimensions are 64. \Cref{tab:hp-values-nfrlbc} contains the values of main hyperparameters used by NF-RLBC. All experiments were done on RTX 3090s, A5000s or A6000s and did not require more than 3 hours each. 

\begin{table}[H]
    \centering
    \caption{\footnotesize \textbf{NF-RLBC:} Hyper-parameters and their values.} 
    \begin{tabular}{l c}
        \toprule
        \text{Hyper-parameter} & \text{Description} \\
        \midrule
        channels & 512 \\
        noise std & 0.1 \\
        actor entropy coefficient ($\lambda$ in \cref{eq:maxent-rl}) & 0.0 \\
        actor BC loss coefficient ($\alpha$ in \cref{eq:maxent-rl}) & \Cref{tab:alpha-nfrlbc} \\
        blocks & 6 \\
        encoder num layers & 4 \\
        rep dims & 512 \\
        \bottomrule
        \end{tabular}
    \label{tab:hp-values-nfrlbc}
\end{table}

\begin{table}[H]
    \centering
    \caption{\footnotesize \textbf{NF-RLBC:} Environment wise actor BC loss coefficient. We use the same hyperparameter tuning procedure as all the baselines~\cite{park2025fql}. See App E.2. of~\citet{park2025fql}} 
    \begin{tabular}{l c}
        \toprule
        \text{Environment Name} & \text{Actor BC loss coefficient} $\alpha$ \\
        \midrule
        puzzle $3\times3$ play & 10.0 \\
        scene play & 10.0 \\
        antmaze large navigation & 1.0 \\
        cube single play & 10.0 \\
        humanoidmaze medium navigation & 1.0 \\
        antsoccer arena navigation & 1.0 \\
        \bottomrule
        \end{tabular}
    \label{tab:alpha-nfrlbc}
\end{table}

\subsection{Unsupervised RL: NF-UGS}
\label{app:cond-implementation-url}
For all unsupervised RL experiments, we use a goal conditioned value function with 6 NF blocks. The fully connected networks of the coupling network for each NF block consist of two hidden layers with 256 activations each. We use a fully connected network for the state-action encoder with 4 hidden layers having 1024 activations each. We train both the conditional $p_{\theta}(g \mid s,a)$ and the unconditional $p_{\theta}(g)$ using the same network. We apply a mask on the conditional variable inputs with probability $0.1$. The discount factor we use is $0.99$. We add a gaussian noise with standard deviation $0.05$. Because we do not sample from the model in NF-UGS, we do not need to employ the denoising trick. Instead, 1024 random goals from the buffer are sampled and the one with the lowest marginal density $p_{\theta}(g)$ is selected for data collection. All experiments were done on RTX 3090s, A5000s or A6000s and did not require more than 20 hours each. 

\section{More details about NF block.}
\label{app:nf-details}
\begin{figure}[h] 
\centering
\includegraphics[width=0.35\textwidth]{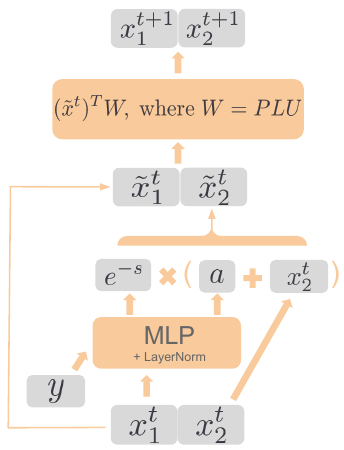}
\caption{\footnotesize \textbf{NF block.}}
\label{fig:nf-block}
\end{figure}

It requires the same cost as the coupling of the forward block to calculate the coupling of the inverse block ${f^{t}_{\theta}}^{-1}$:
\begin{align*}
	&\tilde{x}_1^{t}, \tilde{x}_2^{t} = \text{split}(\tilde{x}^{t}), \\
	&x_1^t = \tilde{x}_1^{t}, \\
	& x_2^{t} = \tilde{x}_2^{t} \times \exp( s_{\theta}^t(x_1^t, y) ) - a_{\theta}^t(x_1^t, y)), \\
	&x^{t} = \text{concat}(x_1^t, x_2^t).
\end{align*}

See \cref{fig:nf-block} for a visualization of each NF block. 

\begin{table}[h]
\centering
\caption{\textbf{Imitation Learning}: Performance with varying numbers of NF blocks on the Push-T task.}
\label{tab:abl_nf_blocks}
\begin{tabular}{cc}
\toprule
\textbf{number of NF blocks} & \textbf{Push-T} \\
\midrule
6  & 0.77  \\
14 & 0.79  \\
24 & 0.827 \\
28 & 0.859 \\
32 & 0.85  \\
40 & 0.844 \\
48 & 0.87  \\
54 & 0.85  \\
\bottomrule
\end{tabular}
\end{table}

\begin{table}[h]
\centering % To center the table
\caption{\textbf{Imitation Learning}: Performance for different NF coupling layer widths on the Push-T task.}
\label{tab:abl_nf_width}
\begin{tabular}{cc}
\toprule
\textbf{width of NF's coupling layers} & {\textbf{Push-T}} \\
\midrule
128  & 0.79  \\
256  & 0.82  \\
1024 & 0.823 \\
\bottomrule
\end{tabular}
\end{table}

\begin{table}[h]
\centering % To center the table
\caption{\textbf{Imitation Learning}: Performance with varying noise levels for the denoising trick~(\cref{app:implementation-il}, ~\citep{zhai2024tarf}) on the Push-T task.}
\label{tab:abl_noise_std}
\begin{tabular}{cc}
\toprule
\textbf{noise\_std} & {\textbf{Push-T}} \\
\midrule
0.5  & 0.77  \\
0.3  & 0.79  \\
0.1  & 0.827 \\
0.05 & 0.84  \\
0.03 & 0.873 \\
0.01 & 0.83  \\
0.0  & {NaN} \\
\bottomrule
\end{tabular}
\end{table}

\begin{table}[h]
\centering
\caption{\textbf{Conditional Imitation Learning:} Ablation study for the importance of various components of NF-GCBC. These experiments show that the architectural modifications we propose are crucial for performance of NFs in the context of RL. We hope that the above experiments serve as sufficient motivation for our design choices.}
\label{tab:ablation_components}
\begin{tabular}{lccc}
\toprule
\textbf{Method} & {\textbf{antmaze-large-navigate}} & {\textbf{scene-play}} & {\textbf{cube-single-play}} \\
\midrule
NF-GCBC                     & 0.32  & 0.42 & 0.795 \\
without permutation flows   & 0.26  & 0.04 & 0.0   \\
without LayerNorm           & 0.29  & 0.3  & 0.0   \\
without sampling trick      & 0.3   & 0.42 & 0.69  \\
\bottomrule
\end{tabular}
\end{table}

\section{Additional results.}
\label{app:additional-results}
\Cref{tab:gcil-extra-result} and \cref{tab:offlinerl-extra-result} contain additional experimental results.

\section{Ablation experiments.}
\label{app:ablation-results}
In~\cref{tab:abl_nf_width}, \cref{tab:abl_nf_blocks}, \cref{tab:abl_noise_std} we ablate the values of the width of the coupling networks, the number of NF blocs, and the levels of noise for the denoising trick~\cref{app:implementation-il} for NF-BC. These ablation experiments aim to show which hyper-parameters are important to tune and which hyper-parameters NF-BC is robust to.

In~\cref{tab:ablation_components}, we present experiments ablating the inclusion of LayerNorm, a generalized permutation layer, and the sampling trick in NF-GCBC.

\begin{table}[h]
\small
\centering
\begin{threeparttable}
\caption{\footnotesize \textbf{Task wise performance for \cref{fig:gcil-result}.}}
\begin{tabular}{l c}
\toprule
\textbf{Environment} & \textbf{Score} \\
\midrule
puzzle-3x3-play-v0 & $0.045\scriptstyle{\pm0.005}$ \\
cube-single-play-v0 & $0.795\scriptstyle{\pm0.109}$ \\
scene-play-v0 & $0.420\scriptstyle{\pm0.008}$ \\
cube-double-play-v0 & $0.351\scriptstyle{\pm0.087}$ \\
antmaze-medium-stitch-v0 & $0.566\scriptstyle{\pm0.068}$ \\
humanoidmaze-medium-navigate-v0 & $0.314\scriptstyle{\pm0.042}$ \\
antmaze-large-navigate-v0 & $0.320\scriptstyle{\pm0.055}$ \\
antsoccer-arena-navigate-v0 & $0.621\scriptstyle{\pm0.007}$ \\
antsoccer-arena-stitch-v0 & $0.643\scriptstyle{\pm0.016}$ \\
\bottomrule
\end{tabular}
\label{tab:gcil-extra-result}
\end{threeparttable}
\vspace{-0.5em}
\end{table}

\begin{table}[h]
\small
\centering
\begin{threeparttable}
\caption{\footnotesize \textbf{Task wise performance for \cref{fig:offlinerl-result}.} Scores are mean $\scriptstyle{\pm}$ std.}
\begin{tabular}{l c}
\toprule
\textbf{Environment} & \textbf{Score} \\
\midrule
\midrule
antmaze-large-navigate-singletask-task1-v0 & $0.764\scriptstyle{\pm0.073}$ \\
antmaze-large-navigate-singletask-task2-v0 & $0.724\scriptstyle{\pm0.088}$ \\
antmaze-large-navigate-singletask-task3-v0 & $0.924\scriptstyle{\pm0.023}$ \\
antmaze-large-navigate-singletask-task4-v0 & $0.616\scriptstyle{\pm0.314}$ \\
antmaze-large-navigate-singletask-task5-v0 & $0.608\scriptstyle{\pm0.399}$ \\
\midrule
antsoccer-arena-navigate-singletask-task1-v0 & $0.600\scriptstyle{\pm0.122}$ \\
antsoccer-arena-navigate-singletask-task2-v0 & $0.556\scriptstyle{\pm0.096}$ \\
antsoccer-arena-navigate-singletask-task3-v0 & $0.420\scriptstyle{\pm0.077}$ \\
antsoccer-arena-navigate-singletask-task4-v0 & $0.480\scriptstyle{\pm0.061}$ \\
antsoccer-arena-navigate-singletask-task5-v0 & $0.336\scriptstyle{\pm0.062}$ \\
\midrule
cube-single-play-singletask-task1-v0 & $0.440\scriptstyle{\pm0.057}$ \\
cube-single-play-singletask-task2-v0 & $0.640\scriptstyle{\pm0.087}$ \\
cube-single-play-singletask-task3-v0 & $0.768\scriptstyle{\pm0.060}$ \\
cube-single-play-singletask-task4-v0 & $0.600\scriptstyle{\pm0.082}$ \\
cube-single-play-singletask-task5-v0 & $0.616\scriptstyle{\pm0.067}$ \\
\midrule
humanoidmaze-medium-navigate-singletask-task1-v0 & $0.036\scriptstyle{\pm0.045}$ \\
humanoidmaze-medium-navigate-singletask-task2-v0 & $0.188\scriptstyle{\pm0.097}$ \\
humanoidmaze-medium-navigate-singletask-task3-v0 & $0.148\scriptstyle{\pm0.072}$ \\
humanoidmaze-medium-navigate-singletask-task4-v0 & $0.016\scriptstyle{\pm0.015}$ \\
humanoidmaze-medium-navigate-singletask-task5-v0 & $0.208\scriptstyle{\pm0.053}$ \\
\midrule
puzzle-3x3-play-singletask-task1-v0 & $0.996\scriptstyle{\pm0.008}$ \\
puzzle-3x3-play-singletask-task2-v0 & $0.996\scriptstyle{\pm0.008}$ \\
puzzle-3x3-play-singletask-task3-v0 & $0.976\scriptstyle{\pm0.029}$ \\
puzzle-3x3-play-singletask-task4-v0 & $0.996\scriptstyle{\pm0.008}$ \\
puzzle-3x3-play-singletask-task5-v0 & $0.996\scriptstyle{\pm0.008}$ \\
\midrule
scene-play-singletask-task1-v0 & $1.000\scriptstyle{\pm0.000}$ \\
scene-play-singletask-task2-v0 & $0.880\scriptstyle{\pm0.075}$ \\
scene-play-singletask-task3-v0 & $0.988\scriptstyle{\pm0.016}$ \\
scene-play-singletask-task4-v0 & $0.856\scriptstyle{\pm0.056}$ \\
scene-play-singletask-task5-v0 & $0.000\scriptstyle{\pm0.000}$ \\
\bottomrule
\end{tabular}
\label{tab:offlinerl-extra-result}
\end{threeparttable}
\vspace{-0.5em}
\end{table}

\newpage

\section{Hyperparameters used by behavior cloning algorithms based on different probabilistic models.}
\label{app:hps-baseline}

\begin{table}[H]
    \centering
    \begin{threeparttable}
    \caption{\footnotesize \textbf{Hyper-parameters used by NF-BC.} } 
    \begin{tabular}{l c}
        \toprule
        \text{Hyper-parameter} & \text{Description} \\
        \midrule
        channels & width of each layer of the coupling MLP network. \\
        noise std & added noise to labels for training the NF using MLE. \\
        blocks & number of NF blocks. \\
        encoder num layers & number of layers in the observation encoder MLP network. \\
        rep dims & output dimension of observation encoder MLP network. \\
        \bottomrule
        \end{tabular}
    \label{tab:hps-nfbc}
    \end{threeparttable}
\end{table}

\begin{table}[H]
    \centering
    \begin{threeparttable}
    \caption{\footnotesize \textbf{Hyper-parameters used by BET.} } 
    \begin{tabular}{l c}
        \toprule
        \text{Hyper-parameter} & \text{Description} \\
        \midrule
        embedding dims & dimension length for attention vectors. \\
        num layers & number of attention blocks. \\
        num heads & number of heads in attention. \\
        act scale & action scaling parameter. \\
        num bins & number of centroids for k-means.\\
        offset prediction & whether to predict offsets for actions or not.\\
        offset loss scale & action offset prediction loss coefficient.\\
        gamma & focal loss parameter.\\
        \bottomrule
        \end{tabular}
    \label{tab:hps-bet}
    \end{threeparttable}
\end{table}

\begin{table}[H]
    \centering
    \begin{threeparttable}
    \caption{\footnotesize \textbf{Hyper-parameters used by DiffPolicy-C.} } 
    \begin{tabular}{l c}
        \toprule
        \text{Hyper-parameter} & \text{Description} \\
        \midrule
        time-step embedding dims & embedding dimension of diffusion timestep. \\
        down dims & downsampling dimensions for Unet architecture. \\
        kernel size & kernel size for CNNs in Unet. \\
        n groups & number of groups for group normalization. \\
        scheduler train timesteps & Length of the forward diffusion chain while training. \\
        num inference steps & Length of the denoising steps while sampling. \\
        beta schedule type & type of scheduler for noising.\\
        beta start & first-step noise for diffusion scheduler. \\
        beta end & last-step noise for diffusion scheduler. \\
        ema inverse gamma & parameter for ema rate schedule. \\
        ema power & parameter for ema rate schedule. \\
        ema min & minimum value of ema decay rate. \\
        ema max & maximum value of ema decay rate. \\
        \bottomrule
        \end{tabular}
    \label{tab:hps-DiffPolicy-C}
    \end{threeparttable}
\end{table}

\begin{table}[H]
    \centering
    \begin{threeparttable}
    \caption{\footnotesize \textbf{Hyper-parameters used by DiffPolicy-T.} } 
    \begin{tabular}{l c}
        \toprule
        \text{Hyper-parameter} & \text{Description} \\
        \midrule
        time-step embedding dims & embedding dimension of diffusion timestep. \\
        embedding dims & dimension length for attention vectors. \\
        num layers & number of attention blocks. \\
        num heads & number of heads in attention. \\
        scheduler train timesteps & Length of the forward diffusion chain while training. \\
        num inference steps & Length of the denoising steps while sampling. \\
        beta schedule type & type of scheduler for noising.\\
        beta start & first-step noise for diffusion scheduler. \\
        beta end & last-step noise for diffusion scheduler. \\
        ema inverse gamma & parameter for ema rate schedule. \\
        ema power & parameter for ema rate schedule. \\
        ema min & minimum value of ema decay rate. \\
        ema max & maximum value of ema decay rate. \\
        \bottomrule
        \end{tabular}
    \label{tab:hps-DiffPolicy-T}
    \end{threeparttable}
\end{table}

\begin{table}[H]
    \centering
    \begin{threeparttable}
    \caption{\footnotesize \textbf{Hyper-parameters used by VQ-BET.} } 
    \begin{tabular}{l c}
        \toprule
        \text{Hyper-parameter} & \text{Description} \\
        \midrule
        embedding dims & dimension length for attention vectors. \\
        num layers & number of attention blocks. \\
        num heads & number of heads in attention. \\
        action seq len & sequence length used to train the behavior transformer. \\
        vqvae num embed & cookbook size of vqvae. \\
        vqvae embedding dims & embedding dimensions of vqvae codes. \\
        vqvae groups & number of groups in residual vqvae. \\
        encoder loss multiplier & encoder loss coefficient.\\
        offset loss multiplier & action offset prediction loss coefficient.\\
        secondary code multiplier & secondary code prediction loss coefficient.\\
        gamma & focal loss parameter.\\
        \bottomrule
        \end{tabular}
    \label{tab:hps-vqbet}
    \end{threeparttable}
\end{table}

\section{Societal Impacts}
\label{app:societal-impacts}
\textbf{Positive Impacts.} Our work directly improves various RL algorithms in various RL problem settings, which can benefit domains like robotics, healthcare, and energy systems by enabling efficient autonomous decision-making.

\textbf{Negative Impacts.} As with many advances in reinforcement learning, improvements in policy expressiveness can be misused in domains such as surveillance, autonomous weapons, or manipulation in recommendation systems. Further, increased training complexity may lead to higher computational costs, raising concerns about energy use and environmental impact.

\section*{NeurIPS Paper Checklist}

\begin{enumerate}

\item {\bf Claims}
    \item[] Question: Do the main claims made in the abstract and introduction accurately reflect the paper's contributions and scope?
    \item[] Answer: \answerYes{} % Replace by \answerYes{}, \answerNo{}, or \answerNA{}.
    \item[] Justification: The main claims made in the abstract and introduction are that our experiments show that NFs are expressive, flexible and generally capable models for many RL problems like imitation learning, offline, goal conditioned RL and unsupervised RL. Towards the end of our introduction, we highlight the success results of NF based algorithms in all of these problem settings. \Cref{sec:experiments} contains detailed discussions of these experiments and results which concretely justify these claims.

\item {\bf Limitations}
    \item[] Question: Does the paper discuss the limitations of the work performed by the authors?
    \item[] Answer: \answerYes{} %, \answerNo{}, or \answerNA{}.
    \item[] Justification: Yes, the second paragraph of \cref{sec:conclusion} discussion the main limitations of our work.

\item {\bf Theory assumptions and proofs}
    \item[] Question: For each theoretical result, does the paper provide the full set of assumptions and a complete (and correct) proof?
    \item[] Answer: \answerNA{} % Replace by \answerYes{}, \answerNo{}, or \answerNA{}.
    \item[] Our paper is primarily an empirical paper. While \cref{sec:rl-vi-mle} does provide a unifying probabilistic framework to think about many RL algorithms, the specific algorithm-wise results are known and have been cited accordingly.

    \item {\bf Experimental result reproducibility}
    \item[] Question: Does the paper fully disclose all the information needed to reproduce the main experimental results of the paper to the extent that it affects the main claims and/or conclusions of the paper (regardless of whether the code and data are provided or not)?
    \item[] Answer: \answerYes{}
    \item[] Yes, the paper provide all information needed to reproduce the all the experimental results. \Cref{app:implementation} contains the implementation details of all the algorithms, including the hyper-parameters used and their values. We have also provided the code for all our experiments organized in a readable manner in supplementary experiments. We have also provided the code for our architecture in both Jax~\cite{jax2018github} and PyTorch~\cite{paszke2019pytorch}.

\item {\bf Open access to data and code}
    \item[] Question: Does the paper provide open access to the data and code, with sufficient instructions to faithfully reproduce the main experimental results, as described in supplemental material?
    \item[] Answer: \answerYes{}
    \item[] Like mentioned in the previous answer, we have provided the code for all our experiments organized in a readable manner in supplementary experiments. We have also provided the code for our architecture in both Jax~\cite{jax2018github} and PyTorch~\cite{paszke2019pytorch}.

\item {\bf Experimental setting/details}
    \item[] Question: Does the paper specify all the training and test details (e.g., data splits, hyperparameters, how they were chosen, type of optimizer, etc.) necessary to understand the results?
    \item[] Answer: \answerYes{} % Replace by \answerYes{}, \answerNo{}, or \answerNA{}.
    \item[] \Cref{app:implementation} contains the implementation details of all the algorithms, including the hyper-parameters used, their values, and details of how they were chosen. Every subsection of~\cref{sec:experiments} contains detailed description of the tasks and baselines for all RL problems (offline imitation learning, conditional offline imitation learning, offline RL, unsupervised goal conditioned RL). 

\item {\bf Experiment statistical significance}
    \item[] Question: Does the paper report error bars suitably and correctly defined or other appropriate information about the statistical significance of the experiments?
    \item[] Answer: \answerYes{} % Replace by \answerYes{}, \answerNo{}, or \answerNA{}.
    \item[] Justification: Yes, all experiments are conducted across five seeds and error bars are shown for all bar / graph plots.

\item {\bf Experiments compute resources}
    \item[] Question: For each experiment, does the paper provide sufficient information on the computer resources (type of compute workers, memory, time of execution) needed to reproduce the experiments?
    \item[] Answer:  \answerYes{}% Replace by \answerYes{}, \answerNo{}, or \answerNA{}.
    \item[] Justification: In \cref{app:implementation}, we provide the type of GPUs used and the time taken for all our experiments. 
    
\item {\bf Code of ethics}
    \item[] Question: Does the research conducted in the paper conform, in every respect, with the NeurIPS Code of Ethics \url{https://neurips.cc/public/EthicsGuidelines}?
    \item[] Answer: \answerYes{} % Replace by \answerYes{}, \answerNo{}, or \answerNA{}.
    \item[] Justification: I have read the guidelines and our research does indeed conform, in every respect, with the NeurIPS Code of Ethics.

\item {\bf Broader impacts}
    \item[] Question: Does the paper discuss both potential positive societal impacts and negative societal impacts of the work performed?
    \item[] Answer: \answerYes{} % Replace by \answerYes{}, \answerNo{}, or \answerNA{}.
    \item[] Justification: In \cref{app:societal-impacts} we have added a brief discussion of both positive and potentially negative societal impacts of our work.
    
\item {\bf Safeguards}
    \item[] Question: Does the paper describe safeguards that have been put in place for responsible release of data or models that have a high risk for misuse (e.g., pretrained language models, image generators, or scraped datasets)?
    \item[] Answer: \answerNA{} % Replace by \answerYes{}, \answerNo{}, or \answerNA{}.
    \item[] Justification: We do not use any of data or models that have a high risk for misuse.

\item {\bf Licenses for existing assets}
    \item[] Question: Are the creators or original owners of assets (e.g., code, data, models), used in the paper, properly credited and are the license and terms of use explicitly mentioned and properly respected?
    \item[] Answer: \answerYes{} % Replace by \answerYes{}, \answerNo{}, or \answerNA{}.
    \item[] Justification: Wherever necessary, we have cited the original source of the datasets and benchmarks used in our paper~\cref{sec:experiments}.

\item {\bf New assets}
    \item[] Question: Are new assets introduced in the paper well documented and is the documentation provided alongside the assets?
    \item[] Answer: \answerYes{} % Replace by \answerYes{}, \answerNo{}, or \answerNA{}.
    \item[] Justification: Yes, our code is well documented. Details about the architecture, experiments and results have been provided in the paper~\cref{sec:experiments},~\cref{app:implementation}.

\item {\bf Crowdsourcing and research with human subjects}
    \item[] Question: For crowdsourcing experiments and research with human subjects, does the paper include the full text of instructions given to participants and screenshots, if applicable, as well as details about compensation (if any)? 
    \item[] Answer: \answerNA{} % Replace by \answerYes{}, \answerNo{}, or \answerNA{}.
    \item[] Justification: There are no crowdsourcing experiments in our paper.

\item {\bf Institutional review board (IRB) approvals or equivalent for research with human subjects}
    \item[] Question: Does the paper describe potential risks incurred by study participants, whether such risks were disclosed to the subjects, and whether Institutional Review Board (IRB) approvals (or an equivalent approval/review based on the requirements of your country or institution) were obtained?
    \item[] Answer: \answerNA{} % Replace by \answerYes{}, \answerNo{}, or \answerNA{}.
    \item[] Justification: the paper does not involve crowdsourcing nor research with human subjects.

\item {\bf Declaration of LLM usage}
    \item[] Question: Does the paper describe the usage of LLMs if it is an important, original, or non-standard component of the core methods in this research? Note that if the LLM is used only for writing, editing, or formatting purposes and does not impact the core methodology, scientific rigorousness, or originality of the research, declaration is not required.
    %this research? 
    \item[] Answer: \answerNA{} % Replace by \answerYes{}, \answerNo{}, or \answerNA{}.
    \item[] Justification: The core method development in this research does not involve LLMs as any important, original, or non-standard components.

\end{enumerate}


\begin{thebibliography}{}

\bibitem[Agarwal et~al., 2020]{agarwal2020opt}
Agarwal, R., Schuurmans, D., and Norouzi, M. (2020).
\newblock An optimistic perspective on offline reinforcement learning.
\newblock In III, H.~D. and Singh, A., editors, {\em Proceedings of the 37th International Conference on Machine Learning}, volume 119 of {\em Proceedings of Machine Learning Research}, pages 104--114. PMLR.

\bibitem[Ajay et~al., 2023]{ajay2023cmayn}
Ajay, A., Du, Y., Gupta, A., Tenenbaum, J., Jaakkola, T., and Agrawal, P. (2023).
\newblock Is conditional generative modeling all you need for decision-making?

\bibitem[Ba et~al., 2016]{ba2016ln}
Ba, J.~L., Kiros, J.~R., and Hinton, G.~E. (2016).
\newblock Layer normalization.

\bibitem[Bain and Sammut, 1995]{Bain1995_bc}
Bain, M. and Sammut, C. (1995).
\newblock A framework for behavioural cloning.
\newblock In {\em Machine Intelligence 15}.

\bibitem[Bengio et~al., 2021]{bengio2021gflow}
Bengio, E., Jain, M., Korablyov, M., Precup, D., and Bengio, Y. (2021).
\newblock Flow network based generative models for non-iterative diverse candidate generation.

\bibitem[Bengio et~al., 2023]{bengio2023gflow}
Bengio, Y., Lahlou, S., Deleu, T., Hu, E.~J., Tiwari, M., and Bengio, E. (2023).
\newblock Gflownet foundations.

\bibitem[Black et~al., 2024]{black2024pi0}
Black, K., Brown, N., Driess, D., Esmail, A., Equi, M., Finn, C., Fusai, N., Groom, L., Hausman, K., Ichter, B., Jakubczak, S., Jones, T., Ke, L., Levine, S., Li-Bell, A., Mothukuri, M., Nair, S., Pertsch, K., Shi, L.~X., Tanner, J., Vuong, Q., Walling, A., Wang, H., and Zhilinsky, U. (2024).
\newblock $\pi_0$: A vision-language-action flow model for general robot control.

\bibitem[Blei et~al., 2017]{Blei_2017}
Blei, D.~M., Kucukelbir, A., and and, J. D.~M. (2017).
\newblock Variational inference: A review for statisticians.
\newblock {\em Journal of the American Statistical Association}, 112(518):859--877.

\bibitem[Bortkiewicz et~al., 2025]{bortkiewicz2024jaxgcrl}
Bortkiewicz, M., Pa{\l}ucki, W., Myers, V., Dziarmaga, T., Arczewski, T., Kuci{\'n}ski, {\L}., and Eysenbach, B. (2025).
\newblock Accelerating goal-conditioned reinforcement learning algorithms and research.
\newblock In {\em The Thirteenth International Conference on Learning Representations}.

\bibitem[Bradbury et~al., 2018]{jax2018github}
Bradbury, J., Frostig, R., Hawkins, P., Johnson, M.~J., Leary, C., Maclaurin, D., Necula, G., Paszke, A., Vander{P}las, J., Wanderman-{M}ilne, S., and Zhang, Q. (2018).
\newblock {JAX}: composable transformations of {P}ython+{N}um{P}y programs.
\newblock GitHub repository.

\bibitem[Brohan et~al., 2023]{brohan2023rt1}
Brohan, A., Brown, N., Carbajal, J., Chebotar, Y., Dabis, J., Finn, C., Gopalakrishnan, K., Hausman, K., Herzog, A., Hsu, J., , et~al. (2023).
\newblock Rt-1: Robotics transformer for real-world control at scale.

\bibitem[Chao et~al., 2024]{chao2024meow}
Chao, C.-H., Feng, C., Sun, W.-F., Lee, C.-K., See, S., and Lee, C.-Y. (2024).
\newblock Maximum entropy reinforcement learning via energy-based normalizing flow.
\newblock In {\em The Thirty-eighth Annual Conference on Neural Information Processing Systems}.

\bibitem[Chebotar et~al., 2023]{chebotar2023qtransf}
Chebotar, Y., Vuong, Q., Hausman, K., Xia, F., Lu, Y., Irpan, A., Kumar, A., Yu, T., Herzog, A., Pertsch, K., Gopalakrishnan, K., Ibarz, J., Nachum, O., Sontakke, S.~A., Salazar, G., Tran, H.~T., Peralta, J., Tan, C., Manjunath, D., Singh, J., Zitkovich, B., Jackson, T., Rao, K., Finn, C., and Levine, S. (2023).
\newblock Q-transformer: Scalable offline reinforcement learning via autoregressive q-functions.
\newblock In Tan, J., Toussaint, M., and Darvish, K., editors, {\em Proceedings of The 7th Conference on Robot Learning}, volume 229 of {\em Proceedings of Machine Learning Research}, pages 3909--3928. PMLR.

\bibitem[Chen et~al., 2021]{chen2021dt}
Chen, L., Lu, K., Rajeswaran, A., Lee, K., Grover, A., Laskin, M., Abbeel, P., Srinivas, A., and Mordatch, I. (2021).
\newblock Decision transformer: Reinforcement learning via sequence modeling.
\newblock In Beygelzimer, A., Dauphin, Y., Liang, P., and Vaughan, J.~W., editors, {\em Advances in Neural Information Processing Systems}.

\bibitem[Chi et~al., 2024]{chi2024diffpol}
Chi, C., Xu, Z., Feng, S., Cousineau, E., Du, Y., Burchfiel, B., Tedrake, R., and Song, S. (2024).
\newblock Diffusion policy: Visuomotor policy learning via action diffusion.

\bibitem[Collaboration et~al., 2024]{openx}
Collaboration, E., O'Neill, A., Rehman, A., Gupta, A., Maddukuri, A., Gupta, A., Padalkar, A., Lee, A., Pooley, A., Gupta, A., Mandlekar, A., et~al. (2024).
\newblock Open x-embodiment: Robotic learning datasets and rt-x models.

\bibitem[Cornish et~al., 2020]{cornish2021rel}
Cornish, R., Caterini, A., Deligiannidis, G., and Doucet, A. (2020).
\newblock Relaxing bijectivity constraints with continuously indexed normalising flows.
\newblock In III, H.~D. and Singh, A., editors, {\em Proceedings of the 37th International Conference on Machine Learning}, volume 119 of {\em Proceedings of Machine Learning Research}, pages 2133--2143. PMLR.

\bibitem[Ding et~al., 2019]{ding2020gcbc}
Ding, Y., Florensa, C., Abbeel, P., and Phielipp, M. (2019).
\newblock Goal-conditioned imitation learning.
\newblock In Wallach, H., Larochelle, H., Beygelzimer, A., d\textquotesingle Alch\'{e}-Buc, F., Fox, E., and Garnett, R., editors, {\em Advances in Neural Information Processing Systems}, volume~32. Curran Associates, Inc.

\bibitem[Dinh et~al., 2014]{dinh2015nice}
Dinh, L., Krueger, D., and Bengio, Y. (2014).
\newblock Nice: Non-linear independent components estimation.
\newblock {\em arXiv: Learning}.

\bibitem[Dinh et~al., 2017]{dinh2017realnvp}
Dinh, L., Sohl-Dickstein, J., and Bengio, S. (2017).
\newblock Density estimation using real {NVP}.
\newblock In {\em International Conference on Learning Representations}.

\bibitem[Draxler et~al., 2024]{draxler2025uni}
Draxler, F., Wahl, S., Schnörr, C., and Köthe, U. (2024).
\newblock {{On the Universality of Volume-Preserving and Coupling-Based Normalizing Flows}}.
\newblock {\em arXiv preprint arXiv:2402.06578}.

\bibitem[Emmons et~al., 2022]{emmons2022rvs}
Emmons, S., Eysenbach, B., Kostrikov, I., and Levine, S. (2022).
\newblock Rvs: What is essential for offline {RL} via supervised learning?
\newblock In {\em International Conference on Learning Representations}.

\bibitem[Eysenbach et~al., 2021]{eysenbach2021clearning}
Eysenbach, B., Salakhutdinov, R., and Levine, S. (2021).
\newblock C-learning: Learning to achieve goals via recursive classification.
\newblock In {\em International Conference on Learning Representations}.

\bibitem[Eysenbach et~al., 2022a]{eysenbach2023ocbc}
Eysenbach, B., Udatha, S., Salakhutdinov, R.~R., and Levine, S. (2022a).
\newblock Imitating past successes can be very suboptimal.
\newblock In Koyejo, S., Mohamed, S., Agarwal, A., Belgrave, D., Cho, K., and Oh, A., editors, {\em Advances in Neural Information Processing Systems}, volume~35, pages 6047--6059. Curran Associates, Inc.

\bibitem[Eysenbach et~al., 2022b]{eysenbach2023crl}
Eysenbach, B., Zhang, T., Levine, S., and Salakhutdinov, R. (2022b).
\newblock Contrastive learning as goal-conditioned reinforcement learning.
\newblock In Oh, A.~H., Agarwal, A., Belgrave, D., and Cho, K., editors, {\em Advances in Neural Information Processing Systems}.

\bibitem[Florensa et~al., 2018]{florensa2018ag}
Florensa, C., Held, D., Geng, X., and Abbeel, P. (2018).
\newblock Automatic goal generation for reinforcement learning agents.
\newblock In Dy, J. and Krause, A., editors, {\em Proceedings of the 35th International Conference on Machine Learning}, volume~80 of {\em Proceedings of Machine Learning Research}, pages 1515--1528. PMLR.

\bibitem[Fu et~al., 2021]{fu2021d4rl}
Fu, J., Kumar, A., Nachum, O., Tucker, G., and Levine, S. (2021).
\newblock D4rl: Datasets for deep data-driven reinforcement learning.

\bibitem[Fujimoto and Gu, 2021]{fujimoto2021minorl}
Fujimoto, S. and Gu, S.~S. (2021).
\newblock A minimalist approach to offline reinforcement learning.
\newblock In Ranzato, M., Beygelzimer, A., Dauphin, Y., Liang, P., and Vaughan, J.~W., editors, {\em Advances in Neural Information Processing Systems}, volume~34, pages 20132--20145. Curran Associates, Inc.

\bibitem[Fujimoto et~al., 2019]{fujimoto2019bcq}
Fujimoto, S., Meger, D., and Precup, D. (2019).
\newblock Off-policy deep reinforcement learning without exploration.
\newblock In Chaudhuri, K. and Salakhutdinov, R., editors, {\em Proceedings of the 36th International Conference on Machine Learning}, volume~97 of {\em Proceedings of Machine Learning Research}, pages 2052--2062. PMLR.

\bibitem[Ghosh et~al., 2019]{ghosh2020gcbc}
Ghosh, D., Gupta, A., Reddy, A., Fu, J., Devin, C., Eysenbach, B., and Levine, S. (2019).
\newblock Learning to reach goals via iterated supervised learning.
\newblock {\em arXiv preprint arXiv:1912.06088}.

\bibitem[Graikos et~al., 2022]{graikos2023diffpnp}
Graikos, A., Malkin, N., Jojic, N., and Samaras, D. (2022).
\newblock Diffusion models as plug-and-play priors.
\newblock In Oh, A.~H., Agarwal, A., Belgrave, D., and Cho, K., editors, {\em Advances in Neural Information Processing Systems}.

\bibitem[Grigsby et~al., 2024]{grigsby2024amago}
Grigsby, J., Fan, L., and Zhu, Y. (2024).
\newblock {AMAGO}: Scalable in-context reinforcement learning for adaptive agents.
\newblock In {\em The Twelfth International Conference on Learning Representations}.

\bibitem[Haarnoja et~al., 2018a]{haarnoja2018lsp}
Haarnoja, T., Hartikainen, K., Abbeel, P., and Levine, S. (2018a).
\newblock Latent space policies for hierarchical reinforcement learning.
\newblock In Dy, J. and Krause, A., editors, {\em Proceedings of the 35th International Conference on Machine Learning}, volume~80 of {\em Proceedings of Machine Learning Research}, pages 1851--1860. PMLR.

\bibitem[Haarnoja et~al., 2018b]{haarnoja2018sac}
Haarnoja, T., Zhou, A., Abbeel, P., and Levine, S. (2018b).
\newblock Soft actor-critic: Off-policy maximum entropy deep reinforcement learning with a stochastic actor.
\newblock In Dy, J. and Krause, A., editors, {\em Proceedings of the 35th International Conference on Machine Learning}, volume~80 of {\em Proceedings of Machine Learning Research}, pages 1861--1870. PMLR.

\bibitem[Hafner et~al., 2019]{hafner2019planet}
Hafner, D., Lillicrap, T., Fischer, I., Villegas, R., Ha, D., Lee, H., and Davidson, J. (2019).
\newblock Learning latent dynamics for planning from pixels.
\newblock In Chaudhuri, K. and Salakhutdinov, R., editors, {\em Proceedings of the 36th International Conference on Machine Learning}, volume~97 of {\em Proceedings of Machine Learning Research}, pages 2555--2565. PMLR.

\bibitem[Hafner et~al., 2024]{hafner2024dreamerv3}
Hafner, D., Pasukonis, J., Ba, J., and Lillicrap, T. (2024).
\newblock Mastering diverse domains through world models.

\bibitem[Hansen-Estruch et~al., 2023]{hansenestruch2023idql}
Hansen-Estruch, P., Kostrikov, I., Janner, M., Kuba, J.~G., and Levine, S. (2023).
\newblock Idql: Implicit q-learning as an actor-critic method with diffusion policies.

\bibitem[Ho et~al., 2020]{ho2020diff}
Ho, J., Jain, A., and Abbeel, P. (2020).
\newblock Denoising diffusion probabilistic models.

\bibitem[Ho and Salimans, 2022]{ho2022cfg}
Ho, J. and Salimans, T. (2022).
\newblock Classifier-free diffusion guidance.

\bibitem[Huang et~al., 2018]{huang2018naf}
Huang, C.-W., Krueger, D., Lacoste, A., and Courville, A. (2018).
\newblock Neural autoregressive flows.
\newblock In Dy, J. and Krause, A., editors, {\em Proceedings of the 35th International Conference on Machine Learning}, volume~80 of {\em Proceedings of Machine Learning Research}, pages 2078--2087. PMLR.

\bibitem[Janner et~al., 2022]{janner2022planningdiff}
Janner, M., Du, Y., Tenenbaum, J., and Levine, S. (2022).
\newblock Planning with diffusion for flexible behavior synthesis.
\newblock In Chaudhuri, K., Jegelka, S., Song, L., Szepesvari, C., Niu, G., and Sabato, S., editors, {\em Proceedings of the 39th International Conference on Machine Learning}, volume 162 of {\em Proceedings of Machine Learning Research}, pages 9902--9915. PMLR.

\bibitem[Janner et~al., 2021]{janner2021dd}
Janner, M., Li, Q., and Levine, S. (2021).
\newblock Offline reinforcement learning as one big sequence modeling problem.
\newblock In Ranzato, M., Beygelzimer, A., Dauphin, Y., Liang, P., and Vaughan, J.~W., editors, {\em Advances in Neural Information Processing Systems}, volume~34, pages 1273--1286. Curran Associates, Inc.

\bibitem[Jordan and Bishop, 2001]{Jordan2001AnIT}
Jordan, M.~I. and Bishop, C.~M. (2001).
\newblock An introduction to graphical models.
\newblock Tutorial material.

\bibitem[Jordan et~al., 1999]{Jordan1999vi}
Jordan, M.~I., Ghahramani, Z., Jaakkola, T.~S., and Saul, L.~K. (1999).
\newblock {\em An introduction to variational methods for graphical models}, page 105–161.
\newblock MIT Press, Cambridge, MA, USA.

\bibitem[Kaelbling, 1993]{Kaelbling1993LearningTA}
Kaelbling, L.~P. (1993).
\newblock Learning to achieve goals.
\newblock In {\em International Joint Conference on Artificial Intelligence}.

\bibitem[Kingma and Dhariwal, 2018]{kingma2018glow}
Kingma, D.~P. and Dhariwal, P. (2018).
\newblock Glow: Generative flow with invertible 1x1 convolutions.
\newblock In Bengio, S., Wallach, H., Larochelle, H., Grauman, K., Cesa-Bianchi, N., and Garnett, R., editors, {\em Advances in Neural Information Processing Systems}, volume~31. Curran Associates, Inc.

\bibitem[Kingma et~al., 2016]{kingma2017iafs}
Kingma, D.~P., Salimans, T., Jozefowicz, R., Chen, X., Sutskever, I., and Welling, M. (2016).
\newblock Improved variational inference with inverse autoregressive flow.
\newblock In Lee, D., Sugiyama, M., Luxburg, U., Guyon, I., and Garnett, R., editors, {\em Advances in Neural Information Processing Systems}, volume~29. Curran Associates, Inc.

\bibitem[Kostrikov et~al., 2021]{kostrikov2021iql}
Kostrikov, I., Nair, A., and Levine, S. (2021).
\newblock Offline reinforcement learning with implicit q-learning.

\bibitem[Kumar et~al., 2019]{kumar2019orl}
Kumar, A., Fu, J., Soh, M., Tucker, G., and Levine, S. (2019).
\newblock Stabilizing off-policy q-learning via bootstrapping error reduction.
\newblock In Wallach, H., Larochelle, H., Beygelzimer, A., d\textquotesingle Alch\'{e}-Buc, F., Fox, E., and Garnett, R., editors, {\em Advances in Neural Information Processing Systems}, volume~32. Curran Associates, Inc.

\bibitem[Kumar et~al., 2020]{kumar2020cql}
Kumar, A., Zhou, A., Tucker, G., and Levine, S. (2020).
\newblock Conservative q-learning for offline reinforcement learning.
\newblock In Larochelle, H., Ranzato, M., Hadsell, R., Balcan, M., and Lin, H., editors, {\em Advances in Neural Information Processing Systems}, volume~33, pages 1179--1191. Curran Associates, Inc.

\bibitem[Laskin et~al., 2020]{srinivas2020curl}
Laskin, M., Srinivas, A., and Abbeel, P. (2020).
\newblock {CURL}: Contrastive unsupervised representations for reinforcement learning.
\newblock In III, H.~D. and Singh, A., editors, {\em Proceedings of the 37th International Conference on Machine Learning}, volume 119 of {\em Proceedings of Machine Learning Research}, pages 5639--5650. PMLR.

\bibitem[Lee et~al., 2020]{lee2020slac}
Lee, A.~X., Nagabandi, A., Abbeel, P., and Levine, S. (2020).
\newblock Stochastic latent actor-critic: Deep reinforcement learning with a latent variable model.
\newblock In Larochelle, H., Ranzato, M., Hadsell, R., Balcan, M., and Lin, H., editors, {\em Advances in Neural Information Processing Systems}, volume~33, pages 741--752. Curran Associates, Inc.

\bibitem[Lee et~al., 2024]{lee2024vqbet}
Lee, S., Wang, Y., Etukuru, H., Kim, H.~J., Shafiullah, N. M.~M., and Pinto, L. (2024).
\newblock Behavior generation with latent actions.
\newblock In Salakhutdinov, R., Kolter, Z., Heller, K., Weller, A., Oliver, N., Scarlett, J., and Berkenkamp, F., editors, {\em Proceedings of the 41st International Conference on Machine Learning}, volume 235 of {\em Proceedings of Machine Learning Research}, pages 26991--27008. PMLR.

\bibitem[Levine, 2018]{levine2018cic}
Levine, S. (2018).
\newblock Reinforcement learning and control as probabilistic inference: Tutorial and review.

\bibitem[Levine et~al., 2020]{levine2020orl}
Levine, S., Kumar, A., Tucker, G., and Fu, J. (2020).
\newblock Offline reinforcement learning: Tutorial, review, and perspectives on open problems.

\bibitem[Lin et~al., 2025]{lin2025datascalinglawsimitation}
Lin, F., Hu, Y., Sheng, P., Wen, C., You, J., and Gao, Y. (2025).
\newblock Data scaling laws in imitation learning for robotic manipulation.

\bibitem[Lipman et~al., 2023]{lipman2023fm}
Lipman, Y., Chen, R. T.~Q., Ben-Hamu, H., Nickel, M., and Le, M. (2023).
\newblock Flow matching for generative modeling.

\bibitem[Lu et~al., 2023]{lu2023synther}
Lu, C., Ball, P.~J., Teh, Y.~W., and Parker-Holder, J. (2023).
\newblock Synthetic experience replay.
\newblock In {\em Thirty-seventh Conference on Neural Information Processing Systems}.

\bibitem[Ma et~al., 2023]{ma2023vipuniversalvisualreward}
Ma, Y.~J., Sodhani, S., Jayaraman, D., Bastani, O., Kumar, V., and Zhang, A. (2023).
\newblock {VIP}: Towards universal visual reward and representation via value-implicit pre-training.
\newblock In {\em The Eleventh International Conference on Learning Representations}.

\bibitem[Micheli et~al., 2023]{micheli2023twm}
Micheli, V., Alonso, E., and Fleuret, F. (2023).
\newblock Transformers are sample-efficient world models.
\newblock In {\em The Eleventh International Conference on Learning Representations}.

\bibitem[Mnih et~al., 2013]{mnih2013dqn}
Mnih, V., Kavukcuoglu, K., Silver, D., Graves, A., Antonoglou, I., Wierstra, D., and Riedmiller, M. (2013).
\newblock Playing atari with deep reinforcement learning.
\newblock {\em arXiv preprint arXiv:1312.5602}.

\bibitem[Myers et~al., 2024]{myers2025tdc}
Myers, V., Zheng, C., Dragan, A.~D., Levine, S., and Eysenbach, B. (2024).
\newblock Learning temporal distances: Contrastive successor features can provide a metric structure for decision-making.
\newblock In {\em ICML}.

\bibitem[Ni et~al., 2022]{ni2022rmfrl}
Ni, T., Eysenbach, B., and Salakhutdinov, R. (2022).
\newblock Recurrent model-free {RL} can be a strong baseline for many {POMDP}s.
\newblock In Chaudhuri, K., Jegelka, S., Song, L., Szepesvari, C., Niu, G., and Sabato, S., editors, {\em Proceedings of the 39th International Conference on Machine Learning}, volume 162 of {\em Proceedings of Machine Learning Research}, pages 16691--16723. PMLR.

\bibitem[OpenAI et~al., 2021]{openai2021asym}
OpenAI, O., Plappert, M., Sampedro, R., Xu, T., Akkaya, I., Kosaraju, V., Welinder, P., D'Sa, R., Petron, A., d.~O.~Pinto, H.~P., Paino, A., Noh, H., Weng, L., Yuan, Q., Chu, C., and Zaremba, W. (2021).
\newblock Asymmetric self-play for automatic goal discovery in robotic manipulation.

\bibitem[Papamakarios et~al., 2021]{papamakarios2021nfs}
Papamakarios, G., Nalisnick, E., Rezende, D.~J., Mohamed, S., and Lakshminarayanan, B. (2021).
\newblock Normalizing flows for probabilistic modeling and inference.
\newblock {\em Journal of Machine Learning Research}, 22(57):1--64.

\bibitem[Papamakarios et~al., 2017]{papamakarios2018maf}
Papamakarios, G., Pavlakou, T., and Murray, I. (2017).
\newblock Masked autoregressive flow for density estimation.
\newblock In Guyon, I., Luxburg, U.~V., Bengio, S., Wallach, H., Fergus, R., Vishwanathan, S., and Garnett, R., editors, {\em Advances in Neural Information Processing Systems}, volume~30. Curran Associates, Inc.

\bibitem[Park et~al., 2025a]{park2025ogbench}
Park, S., Frans, K., Eysenbach, B., and Levine, S. (2025a).
\newblock {OGB}ench: Benchmarking offline goal-conditioned {RL}.
\newblock In {\em The Thirteenth International Conference on Learning Representations}.

\bibitem[Park et~al., 2025b]{park2025fql}
Park, S., Li, Q., and Levine, S. (2025b).
\newblock Flow q-learning.
\newblock {\em ArXiv}.

\bibitem[Paszke et~al., 2019]{paszke2019pytorch}
Paszke, A., Gross, S., Massa, F., Lerer, A., Bradbury, J., Chanan, G., Killeen, T., Lin, Z., Gimelshein, N., Antiga, L., Desmaison, A., Köpf, A., Yang, E., DeVito, Z., Raison, M., Tejani, A., Chilamkurthy, S., Steiner, B., Fang, L., Bai, J., and Chintala, S. (2019).
\newblock Pytorch: An imperative style, high-performance deep learning library.

\bibitem[Pitis et~al., 2020]{pitis2020goalkde}
Pitis, S., Chan, H., Zhao, S., Stadie, B., and Ba, J. (2020).
\newblock Maximum entropy gain exploration for long horizon multi-goal reinforcement learning.
\newblock In III, H.~D. and Singh, A., editors, {\em Proceedings of the 37th International Conference on Machine Learning}, volume 119 of {\em Proceedings of Machine Learning Research}, pages 7750--7761. PMLR.

\bibitem[Pomerleau, 1988]{Pomerleau_bc}
Pomerleau, D.~A. (1988).
\newblock Alvinn: An autonomous land vehicle in a neural network.
\newblock In Touretzky, D., editor, {\em Advances in Neural Information Processing Systems}, volume~1. Morgan-Kaufmann.

\bibitem[Pong et~al., 2020]{pong2020skewfits}
Pong, V.~H., Dalal, M., Lin, S., Nair, A., Bahl, S., and Levine, S. (2020).
\newblock Skew-fit: State-covering self-supervised reinforcement learning.

\bibitem[Ren et~al., 2024]{ren2024dppo}
Ren, A.~Z., Lidard, J., Ankile, L.~L., Simeonov, A., Agrawal, P., Majumdar, A., Burchfiel, B., Dai, H., and Simchowitz, M. (2024).
\newblock Diffusion policy policy optimization.

\bibitem[Rezende and Mohamed, 2015]{rezende2016nf}
Rezende, D. and Mohamed, S. (2015).
\newblock Variational inference with normalizing flows.
\newblock In Bach, F. and Blei, D., editors, {\em Proceedings of the 32nd International Conference on Machine Learning}, volume~37 of {\em Proceedings of Machine Learning Research}, pages 1530--1538, Lille, France. PMLR.

\bibitem[Sermanet et~al., 2018]{sermanet2018tmw}
Sermanet, P., Lynch, C., Chebotar, Y., Hsu, J., Jang, E., Schaal, S., Levine, S., and Brain, G. (2018).
\newblock Time-contrastive networks: Self-supervised learning from video.
\newblock In {\em 2018 IEEE International Conference on Robotics and Automation (ICRA)}, pages 1134--1141.

\bibitem[Shafiullah et~al., 2022]{shafiullah2022cloningk}
Shafiullah, N. M.~M., Cui, Z.~J., Altanzaya, A., and Pinto, L. (2022).
\newblock Behavior transformers: Cloning \$k\$ modes with one stone.
\newblock In Oh, A.~H., Agarwal, A., Belgrave, D., and Cho, K., editors, {\em Advances in Neural Information Processing Systems}.

\bibitem[Singh et~al., 2021]{singh2020parrot}
Singh, A., Liu, H., Zhou, G., Yu, A., Rhinehart, N., and Levine, S. (2021).
\newblock Parrot: Data-driven behavioral priors for reinforcement learning.
\newblock In {\em International Conference on Learning Representations}.

\bibitem[Sohl-Dickstein et~al., 2015]{sohldickstein2015diff}
Sohl-Dickstein, J., Weiss, E.~A., Maheswaranathan, N., and Ganguli, S. (2015).
\newblock Deep unsupervised learning using nonequilibrium thermodynamics.

\bibitem[Song et~al., 2023]{song2023consistencymodels}
Song, Y., Dhariwal, P., Chen, M., and Sutskever, I. (2023).
\newblock Consistency models.

\bibitem[Song et~al., 2021]{song2021diff}
Song, Y., Sohl-Dickstein, J., Kingma, D.~P., Kumar, A., Ermon, S., and Poole, B. (2021).
\newblock Score-based generative modeling through stochastic differential equations.

\bibitem[Stooke et~al., 2021]{pmlr-v139-stooke21a}
Stooke, A., Lee, K., Abbeel, P., and Laskin, M. (2021).
\newblock Decoupling representation learning from reinforcement learning.
\newblock In Meila, M. and Zhang, T., editors, {\em Proceedings of the 38th International Conference on Machine Learning}, volume 139 of {\em Proceedings of Machine Learning Research}, pages 9870--9879. PMLR.

\bibitem[Sukhbaatar et~al., 2018]{sukhbaatar2018imself}
Sukhbaatar, S., Lin, Z., Kostrikov, I., Synnaeve, G., Szlam, A., and Fergus, R. (2018).
\newblock Intrinsic motivation and automatic curricula via asymmetric self-play.

\bibitem[Tabak and Vanden-Eijnden, 2010]{tabak2010nf}
Tabak, E. and Vanden-Eijnden, E. (2010).
\newblock Density estimation by dual ascent of the log-likelihood.
\newblock {\em Communications in Mathematical Sciences}, 8(1):217--233.

\bibitem[Tarasov et~al., 2023]{rebrac}
Tarasov, D., Kurenkov, V., Nikulin, A., and Kolesnikov, S. (2023).
\newblock Revisiting the minimalist approach to offline reinforcement learning.

\bibitem[van~den Oord et~al., 2017]{oord2018vqvae}
van~den Oord, A., Vinyals, O., and kavukcuoglu, k. (2017).
\newblock Neural discrete representation learning.
\newblock In Guyon, I., Luxburg, U.~V., Bengio, S., Wallach, H., Fergus, R., Vishwanathan, S., and Garnett, R., editors, {\em Advances in Neural Information Processing Systems}, volume~30. Curran Associates, Inc.

\bibitem[Vaswani et~al., 2017]{vaswani2023aiayn}
Vaswani, A., Shazeer, N., Parmar, N., Uszkoreit, J., Jones, L., Gomez, A.~N., Kaiser, L.~u., and Polosukhin, I. (2017).
\newblock Attention is all you need.
\newblock In Guyon, I., Luxburg, U.~V., Bengio, S., Wallach, H., Fergus, R., Vishwanathan, S., and Garnett, R., editors, {\em Advances in Neural Information Processing Systems}, volume~30. Curran Associates, Inc.

\bibitem[Venkatraman et~al., 2024]{venkatraman2023ldcq}
Venkatraman, S., Khaitan, S., Akella, R.~T., Dolan, J., Schneider, J., and Berseth, G. (2024).
\newblock Reasoning with latent diffusion in offline reinforcement learning.
\newblock In {\em The Twelfth International Conference on Learning Representations}.

\bibitem[Wang et~al., 2023]{wang2023qrl}
Wang, T., Torralba, A., Isola, P., and Zhang, A. (2023).
\newblock Optimal goal-reaching reinforcement learning via quasimetric learning.
\newblock In Krause, A., Brunskill, E., Cho, K., Engelhardt, B., Sabato, S., and Scarlett, J., editors, {\em Proceedings of the 40th International Conference on Machine Learning}, volume 202 of {\em Proceedings of Machine Learning Research}, pages 36411--36430. PMLR.

\bibitem[Weng et~al., 2022]{pmlr-v164-weng22a}
Weng, T., Bajracharya, S.~M., Wang, Y., Agrawal, K., and Held, D. (2022).
\newblock Fabricflownet: Bimanual cloth manipulation with a flow-based policy.
\newblock In Faust, A., Hsu, D., and Neumann, G., editors, {\em Proceedings of the 5th Conference on Robot Learning}, volume 164 of {\em Proceedings of Machine Learning Research}, pages 192--202. PMLR.

\bibitem[Yarats et~al., 2021]{Yarats_isac}
Yarats, D., Zhang, A., Kostrikov, I., Amos, B., Pineau, J., and Fergus, R. (2021).
\newblock Improving sample efficiency in model-free reinforcement learning from images.
\newblock {\em Proceedings of the AAAI Conference on Artificial Intelligence}, 35(12):10674--10681.

\bibitem[Zhai et~al., 2024]{zhai2024tarf}
Zhai, S., Zhang, R., Nakkiran, P., Berthelot, D., Gu, J., Zheng, H., Chen, T., Bautista, M.~A., Jaitly, N., and Susskind, J. (2024).
\newblock Normalizing flows are capable generative models.
\newblock {\em arXiv preprint arXiv:2412.06329}.

\bibitem[Zheng et~al., 2023]{zheng2023guidedflows}
Zheng, Q., Le, M., Shaul, N., Lipman, Y., Grover, A., and Chen, R.~T. (2023).
\newblock Guided flows for generative modeling and decision making.
\newblock {\em arXiv preprint arXiv:2311.13443}.

\bibitem[Łukasz Kaiser et~al., 2020]{kaiser2024simple}
Łukasz Kaiser, Babaeizadeh, M., Miłos, P., Osiński, B., Campbell, R.~H., Czechowski, K., Erhan, D., Finn, C., Kozakowski, P., Levine, S., Mohiuddin, A., Sepassi, R., Tucker, G., and Michalewski, H. (2020).
\newblock Model based reinforcement learning for atari.
\newblock In {\em International Conference on Learning Representations}.

\end{thebibliography}
\end{document}